\documentclass{article}

\PassOptionsToPackage{numbers, compress, sort}{natbib}
\usepackage[preprint]{neurips_2025}
\usepackage{neurips_2025}
\usepackage{wrapfig}
\usepackage{authblk}

\usepackage[utf8]{inputenc} % allow utf-8 input
\usepackage[T1]{fontenc}    % use 8-bit T1 fonts
\usepackage{hyperref}       % hyperlinks
\usepackage{url}            % simple URL typesetting
\usepackage{booktabs}       % professional-quality tables
\usepackage{amsfonts}       % blackboard math symbols
\usepackage{nicefrac}       % compact symbols for 1/2, etc.
\usepackage{microtype}      % microtypography
         
\usepackage{graphicx}
\usepackage{subfigure}
\usepackage{marvosym}

\title{Prolonged Reasoning Is Not All You Need: Certainty-Based Adaptive Routing for Efficient LLM/MLLM Reasoning}

\author{
    Jinghui Lu \textsuperscript{\rm1}$^{*}$,
    Haiyang Yu \textsuperscript{\rm2}$^{*}$, 
    Siliang Xu \textsuperscript{\rm1}$^{*}$,
    Shiwei Ran \textsuperscript{\rm1}, 
    GuoZhi Tang \textsuperscript{\rm1}, 
    Siqi Wang \textsuperscript{\rm1}, 
    Bin Shan \textsuperscript{\rm1}, 
    Teng Fu \textsuperscript{\rm2}, 
    Hao Feng \textsuperscript{\rm1}, 
    Jingqun Tang \textsuperscript{\rm1}, 
    Han Wang \textsuperscript{\rm1},
    Can Huang \textsuperscript{\rm 1 \Letter}
    \\
    \textsuperscript{1} ByteDance \\
    \textsuperscript{2} FuDan University, China \\
    \texttt{\{lujinghui,xusiliang,ranshiwei,tangguozhi.997\}@bytedance.com} \\
    \texttt{\{wangsiqi.wsq547,shanbin,tangjingqun,wanghan.99,can.huang\}@bytedance.com}
    \texttt{\{hyyu,tfu23\}@fudan.edu.cn}
    \texttt{}\\
}

\begin{document}

\maketitle
\renewcommand{\thefootnote}{}
\footnotetext{$^{*}$ Equal contribution. \Letter ~Corresponding authors.}

\begin{abstract}
Recent advancements in reasoning have significantly enhanced the capabilities of Large Language Models (LLMs) and Multimodal Large Language Models (MLLMs) across diverse tasks. However, excessive reliance on chain-of-thought (CoT) reasoning can impair model performance and brings unnecessarily lengthened outputs, reducing efficiency. Our work reveals that prolonged reasoning does not universally improve accuracy and even degrade performance on simpler tasks. To address this, we propose \textbf{C}ertainty-based \textbf{A}daptive \textbf{R}easoning (CAR), a novel framework that dynamically switches between short answers and long-form reasoning based on the model’s perplexity. CAR first generates a short answer and evaluates its perplexity, triggering reasoning only when the model exhibits low confidence (\textit{i.e.,} high perplexity). Experiments across diverse multimodal VQA/KIE benchmarks and text reasoning datasets show that CAR outperforms both short-answer and long-form reasoning approaches, striking an optimal balance between accuracy and efficiency.
\end{abstract}

\section{Introduction}\label{sec:intro}

Chain-of-thought (CoT)~\cite{kojima2022large,wei2022chain,shah2024causal,lyu2023faithful,zhang2024chain,feng2023towards} reasoning was first introduced to improve the performance of large language models (LLMs) on complex reasoning tasks. By prompting \textit{``Let's think step by step''}, LLMs generates intermediate reasoning steps before outputting the final answer, leading to higher performance compared to directly outputting the answer without such reasoning. Recent reasoning models~\cite{chen2024unlocking,openai_o1,lu2024deepseek,shao2024deepseekmath,guo2025deepseek,meng2024simpo,zhang2023h2o,qwen_qwq,seed2025seed,deepmind2024gemini25,fei2025advancingsequentialnumericalprediction} incorporate CoT reasoning into their training procedures, enabling reasoning without requiring explicit prompting. These reasoning models yield substantial performance improvements especially in tasks with well-structured and deterministic answers, such as mathematics and coding for LLMs~\cite{sprague2025to}, as well as object detection and counting for Multimodal Large Language Models (MLLMs)~\cite{chen2025r1v,shen2025vlm,lu2023punifiedner,lu-etal-2023-makes,lu2024padellm,lu2024bounding,wang2025vision,yu2025eve}.

However, recent studies~\cite{sui2025stop,sprague2025to,feng2025efficient} indicate that long-form reasoning primarily enhances LLM performance on math and symbolic tasks, but offers marginal gains on other types of tasks. \citet{sui2025stop} demonstrate that ``overthinking'' can even hurt model performance in some simpler tasks, highlighting the need for selective application of CoT reasoning. In other words, current models commonly suffer from a ``one-size-fits-all'' application of reasoning strategies. Regardless of the complexity of individual questions, they uniformly employ a long-form reasoning process, resulting inefficient inference and failing to meet real-world demands for both speed and accuracy.

To examine this further, we have performed a preliminary experiment using Qwen2.5-0.5B on three text-rich Visual Question Answering (VQA) and Key Information Extraction (KIE) datasets, where answers are mostly extractive or fact-based. We compare the performance of reasoning-based models against short-answer models (see Appendix~\ref{app:intro_exp} for experimental details), with the results summarized in Table~\ref{tab:preliminary_exp0}. Experimental results support the finding that reasoning does not universally improve performance as evidenced by the higher accuracy of short-answer models (45.7 vs. 34.1). Visualizations (see Appendix~\ref{app:intro_vis}) demonstrate that for certain questions, prolonged reasoning steps may introduce noise, degrading accuracy compared to concise, direct answers. More critically, the high token consumption required for reasoning imposes substantial practical costs. As shown in Table~\ref{tab:preliminary_exp0}, responses with reasoning average 572.3 tokens compared to just 13.9 for direct answers. This inefficiency leads to increased computational overhead, slower inference times, and higher monetary and energy expenses without commensurate accuracy gains in tasks where concise answers suffice.

\begin{table*}[!t]
\centering
% \resizebox{1.0\textwidth}{!}{%
\small
\begin{tabular}{l|cc|cc|cc|cc}
\toprule
~ & \multicolumn{2}{c|}{\textbf{DocVQA}} & \multicolumn{2}{c|}{\textbf{ChartQA}} & \multicolumn{2}{c|}{\textbf{FUNSD}} & \multicolumn{2}{c}{\textbf{Avg.}} \\  \midrule
\textbf{Model} & Acc & \#Token & Acc & \#Token & Acc & \#Token & Acc & \#Token \\
Qwen2.5-0.5B\textsubscript{Short}  & \textbf{53.9} & \textbf{11.6} & \textbf{25.3} & \textbf{13.9} & \textbf{58.0} & \textbf{16.2} & \textbf{45.7} & \textbf{13.9} \\
Qwen2.5-0.5B\textsubscript{Long}  & 41.1 & 505.7 & 16.6 & 727.5 & 44.5 & 483.6 & 34.1 & 572.3 \\
\bottomrule
\end{tabular}%
% }
\caption{Performance comparison of reasoning-based vs. short-answer models using Qwen2.5-0.5B.}
\label{tab:preliminary_exp0}
\end{table*}

This raises the research question:\textit{``Can LLM/MLLM adaptively determine when to engage in reasoning, selectively activating it for questions where simple answers are insufficient to balance computational efficiency and task performance?''} To address this challenge, we propose \textbf{C}ertainty-based \textbf{A}daptive \textbf{R}easoning (CAR), a novel framework that adaptively switches between generating short answers and long-form reasoning based on the model’s confidence.

Specifically, CAR first produces a short answer and evaluates its confidence using perplexity (PPL), a measure of the model’s uncertainty in its response. A high PPL score indicates low confidence. Specifically, we model the relationship between PPL and answer correctness using a Gaussian distribution. For each new short response, we compute the likelihood of its PPL under both Gaussians---if the probability of being incorrect (\textit{i.e.,} PPL’s likelihood under the ``incorrect'' Gaussian) exceeds that of being correct, CAR triggers a full reasoning process to refine the answer. This adaptive approach ensures computational efficiency without sacrificing performance, activating complex reasoning only when necessary.

In this work, we begin with a pilot study to examine the relationship between model confidence and the correctness of model responses, validating the feasibility of using PPL score as an indicator of prediction confidence (Section~\ref{sec:pilot_study}). Building on these observations, we propose Certainty-based Adaptive Reasoning, a method that dynamically routes LLM/MLLM to either short-answer or CoT reasoning (Section~\ref{sec:method}).Our experiments (Section~\ref{sec:main_exp}) demonstrate that CAR achieves higher prediction accuracy compared to both complete short-answer and complete long-form reasoning approaches, while maintaining greater inference efficiency than long-form reasoning across both LLMs and MLLMs. Notably, CAR not only excels in simple tasks such as VQA and KIE but also outperforms conventional reasoning methods in challenging tasks (\textit{e.g.,} mathematics) where long-form reasoning is typically assumed to be necessary.

In summary, the contributions of this paper are as follows:

\begin{itemize}
\item We propose CAR, a novel framework that dynamically switches between concise answers and long-form reasoning based on model confidence, achieving an optimal balance between accuracy and computational efficiency.
\item Through comprehensive pilot studies, we validate PPL can be a reliable indicator of model confidence, establishing its relationship with answer correctness via Gaussian modeling.
\item Extensive experiments across LLMs and MLLMs demonstrate CAR's superiority, showing consistently accuracy improvement over short-answer and long-form reasoning counterparts while significantly reduce the inference tokens.
\item Our experiments demonstrate that CAR surpasses previous state-of-the-art reasoning token reduction methods. When tested with Qwen2.5, CAR improves overall accuracy by 6.9\% while reducing token consumption by 21.4\%. Similarly, with Llama3.1, it achieves a 5.5\% accuracy gain and a 39.0\% reduction in tokens used.
\end{itemize}

To the best of our knowledge, CAR stands as a pioneering approach in adaptive reasoning that automatically determines when to employ reasoning based on real-time confidence estimation, offering a principled solution to the efficiency-accuracy trade-off in LLM/MLLM reasoning.

\section{Related Work}

\subsection{Reasoning Frameworks for LLM/MLLM} Recent advances in reasoning models, such as OpenAI’s o1~\cite{openai_o1}, DeepSeek R1~\cite{guo2025deepseek,shao2024deepseekmath} and QwQ~\cite{qwen_qwq}, have significantly enhanced the ability of LLMs to solve complex tasks, especially in tasks with well-structured and deterministic answers. Chain-of-Thought (CoT) prompting~\cite{kojima2022large,wei2022chain,shah2024causal,lyu2023faithful,zhang2024chain,feng2023towards} is a foundational approach that decomposes problems into intermediate steps, improving accuracy and interpretability. To further improve reasoning performance, researchers have developed various enhancements to CoT. One line of work focuses on aggregating diverse reasoning trajectories to improve reliability, such as sampling multiple reasoning paths and selecting the most consistent answer~\cite{wang2022self,lightman2024lets,kim2023language}. Another direction~\cite{yao2023tree,chen2024boosting,NEURIPS2023_65a39213} explores hierarchical reasoning structures, where models dynamically branch and backtrack to solve multi-step problems more effectively. Some approaches also incorporate external tools and APIs into the reasoning process, enabling models to retrieve information or perform computations beyond their parametric knowledge~\cite{yao2023react,NEURIPS2023_9cb2a749,NEURIPS2023_1b44b878,NEURIPS2024_e4c61f57,NEURIPS2023_a3621ee9}. These innovations have significantly broadened the applicability of reasoning-enhanced LLMs in complex real-world scenarios.

\subsection{Token Reduction Techniques in Reasoning} To the best of our knowledge, CAR is the first work that adaptively decides whether to activate long-form reasoning. The closest related work in the literature is reasoning token reduction methods, which aim to address the high token costs in reasoning inference~\cite{lee2025well,ma2025reasoning}. Several approaches have been proposed to mitigate this issue, each with distinct trade-offs. Concise Thoughts~\cite{nayab2024concise} employs a fixed global token budget, while Token-Budget-Aware LLM Reasoning (TALE)~\cite{han2024token} dynamically adjusts budgets based on problem complexity—though these methods may introduce additional LLM calls or face impractical budget constraints. Also, Chain of Draft (CoD)~\cite{xu2025chain} reduces verbosity by generating minimal intermediate steps, significantly lowering token usage without compromising accuracy. Other techniques include Skeleton-of-Thought~\cite{ning2023skeleton}, which parallelizes reasoning for efficiency, and latent-space reasoning~\cite{hao2024training,shen2025codi,ruan2025reasoning}, which sacrifices interpretability for speed. 

\section{Pilot Study}~\label{sec:pilot_study}

In this section, we conduct a series of exploratory experiments to assess whether PPL can serve as a reliable indicator of model confidence.

% First, we conduct experiments to demonstrate a negative correlation between prediction accuracy and the PPL of short answers at the dataset level—that is, higher PPL (indicating lower confidence in the model's prediction) corresponds to lower accuracy.

% Next, we further validate this relationship by showing that, within each dataset, correctly predicted examples tend to have lower PPL scores than incorrectly predicted ones. Given this negative correlation between PPL and prediction correctness, we propose an empirical method: by thresholding PPL, we can determine whether a prediction is confident enough or requires further reasoning. Specifically, if the PPL exceeds a certain threshold (suggesting low confidence), the model continues processing the question with long-form reasoning.

% Theoretically, if this approach leads to higher prediction accuracy, it would demonstrate the usefulness of PPL as a confidence estimator for model outputs. 

\begin{figure*}[t]
    \centering
    \subfigure[DocVQA]{\includegraphics[width=.24\textwidth]{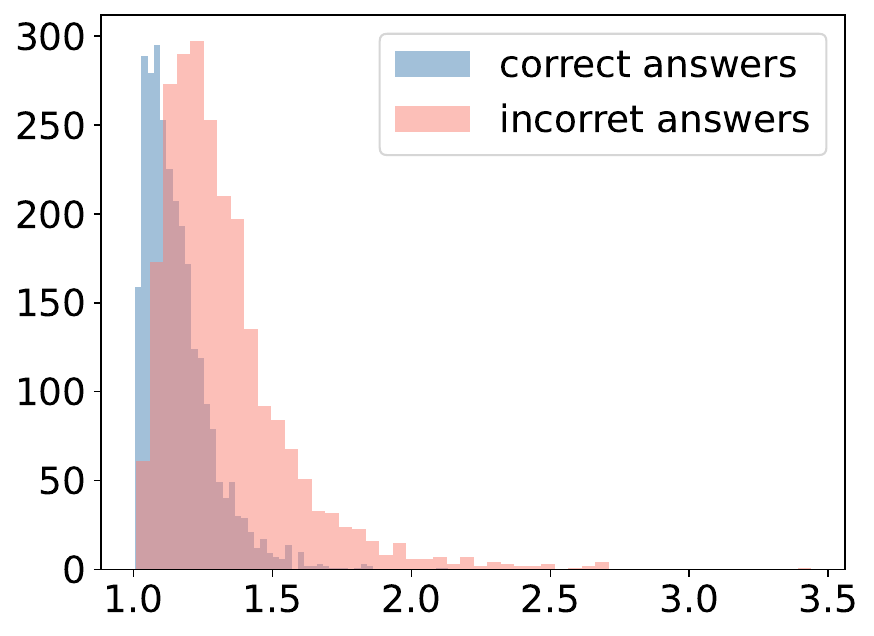}}
    \subfigure[ChartQA]{\includegraphics[width=.24\textwidth]{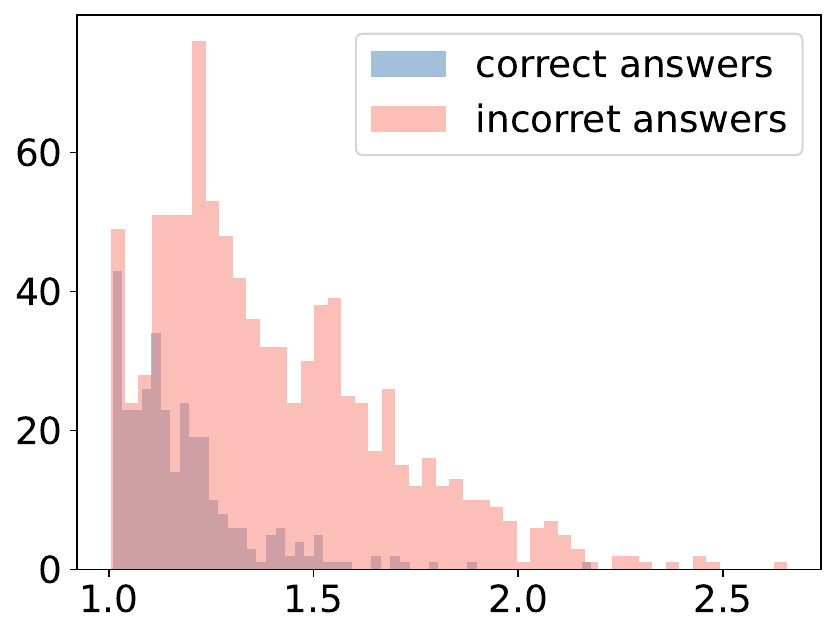}}
    \subfigure[InfoVQA(OOD)]{\includegraphics[width=.24\textwidth]{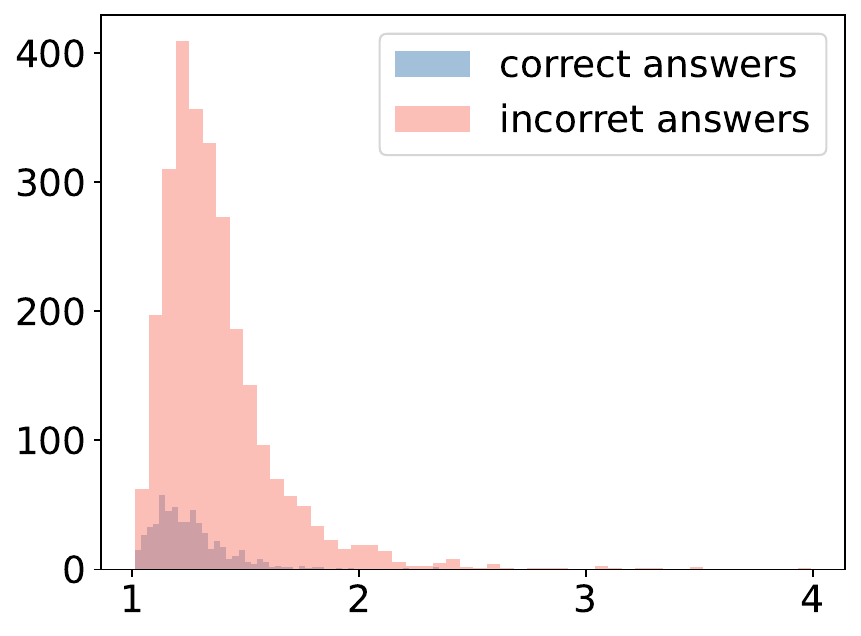}}
    \subfigure[VisualMRC(OOD)]{\includegraphics[width=.24\textwidth]{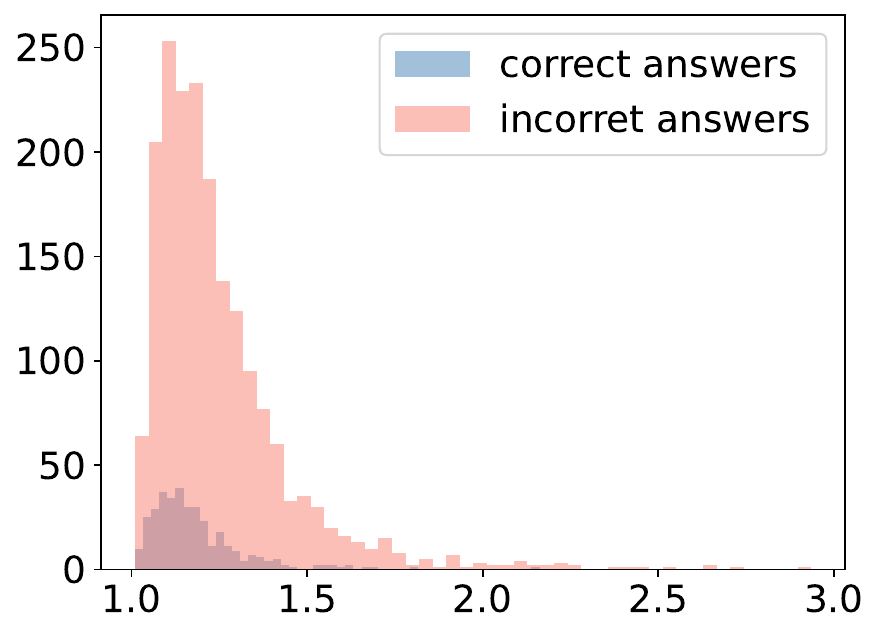}}
    \subfigure[FUNSD]{\includegraphics[width=.24\textwidth]{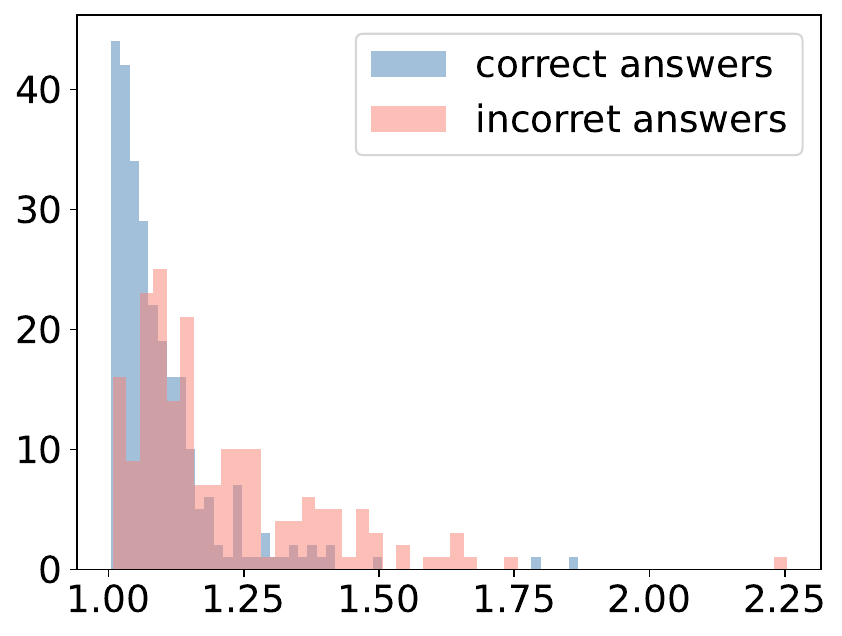}}
    \subfigure[POIE(OOD)]{\includegraphics[width=.24\textwidth]{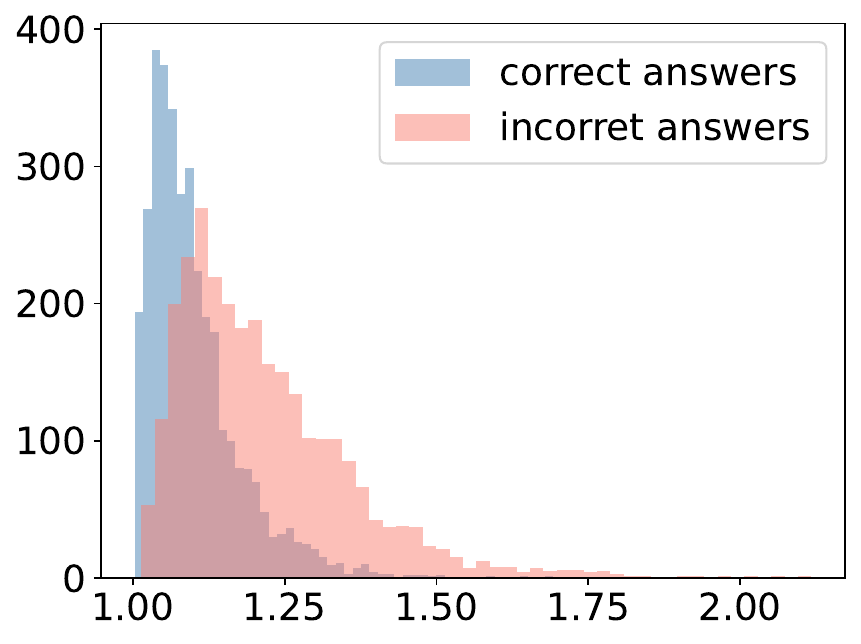}}
    \subfigure[CORD(OOD)]{\includegraphics[width=.24\textwidth]{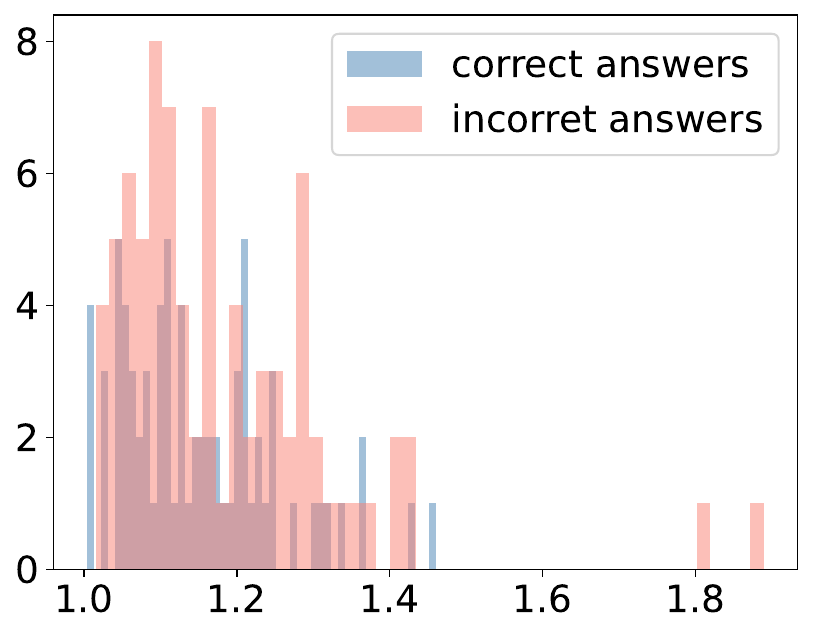}}
    \subfigure[SROIE]{\includegraphics[width=.24\textwidth]{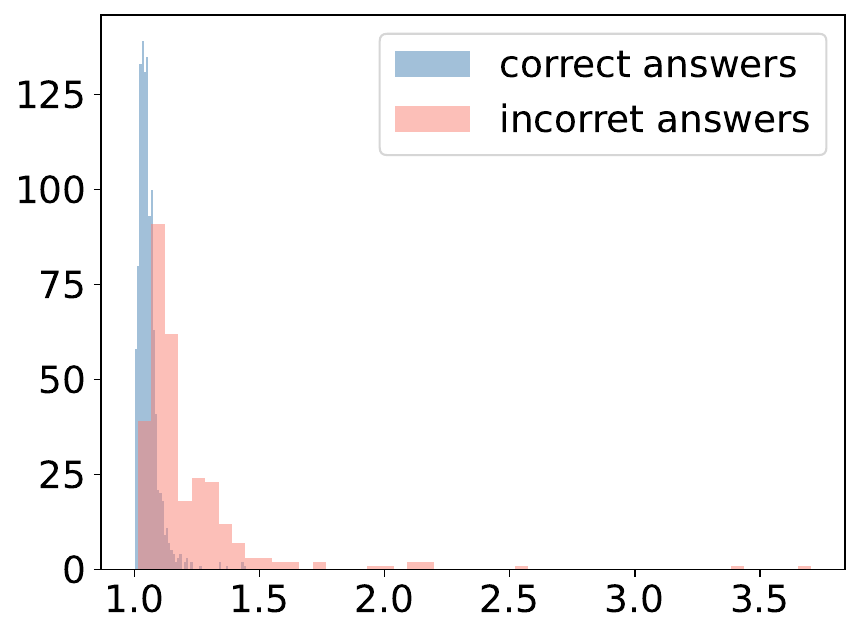}}
    \caption{PPL distribution of Correct and Incorrect Predictions. X-axis is PPL, Y-axis is frequency.}
    \label{fig:good_bad_ppl_05b}
\end{figure*}

\subsection{Datasets} For pilot study, we conduct the investigation on four text-rich VQA datasets \textbf{DocVQA}~\citep{mathew2021docvqa}, \textbf{InfoVQA}~\citep{mathew2022infographicvqa}, \textbf{ChartQA}~\citep{masry-etal-2022-chartqa} and \textbf{VisualMRC}~\citep{tanaka2021visualmrc} and four KIE datasets \textbf{SROIE}~\citep{huang2019icdar2019}, \textbf{CORD}~\citep{park2019cord}, \textbf{FUNSD}~\citep{jaume2019funsd} and \textbf{POIE}~\citep{yang2023modeling}. These datasets are selected since they offer straightforward tasks. Since these datasets are multimodal but the pilot study focuses on LLM-based reasoning, we structure the input using the OCR text and bounding box provided by respective dataset. For example: ``Given document <document>Saint[466,77,554,107]\textbackslash nLouis[561,77,657,107]\textbackslash nHeart[664,77,760,106] ...''. However, these datasets do not include reasoning steps. To address this, we employ DeepSeek-R1 to generate plausible reasoning processes by providing the question and final answer. 

\subsection{Setup} 

We fine-tune Qwen2.5-0.5B on DocVQA, ChartQA, FUNSD, and SROIE, evaluating performance on these in-domain datasets as well as on out-of-domain (OOD) test sets (POIE, InfoVQA, VisualMRC, and CORD). For each example, we train the models to generate two response types: concise short answers when prompted with \textit{``Please directly output the answer''} and long-form reasoning followed by answers when prompted with \textit{``Please output the reasoning process before outputting the answer''}. Additional details can be found in Appendix~\ref{app:exp_setup}. We also fine-tune Qwen2.5-7B using the same settings; the results are provided in Appendix~\ref{app:7b_result} for brevity. The computation of PPL is defined in Equation~\ref{eq:ppl}. All PPL scores are calculated based on the short answer. For model performance, we adopt the accuracy metric as in Section~\ref{subsec:setup}, where a response is considered correct if it encompasses the ground truth.

\subsection{Analysis} 

% In Visual Question Answering (VQA) and Key Information Extraction (KIE), reasoning-enhanced models have become pivotal technologies for improving task performance due to their powerful logical deduction capabilities. 

\begin{wrapfigure}{!t}{0.5\textwidth}  % 'r' for right side and width of the wrap figure
  \centering
  \includegraphics[width=.5\columnwidth]{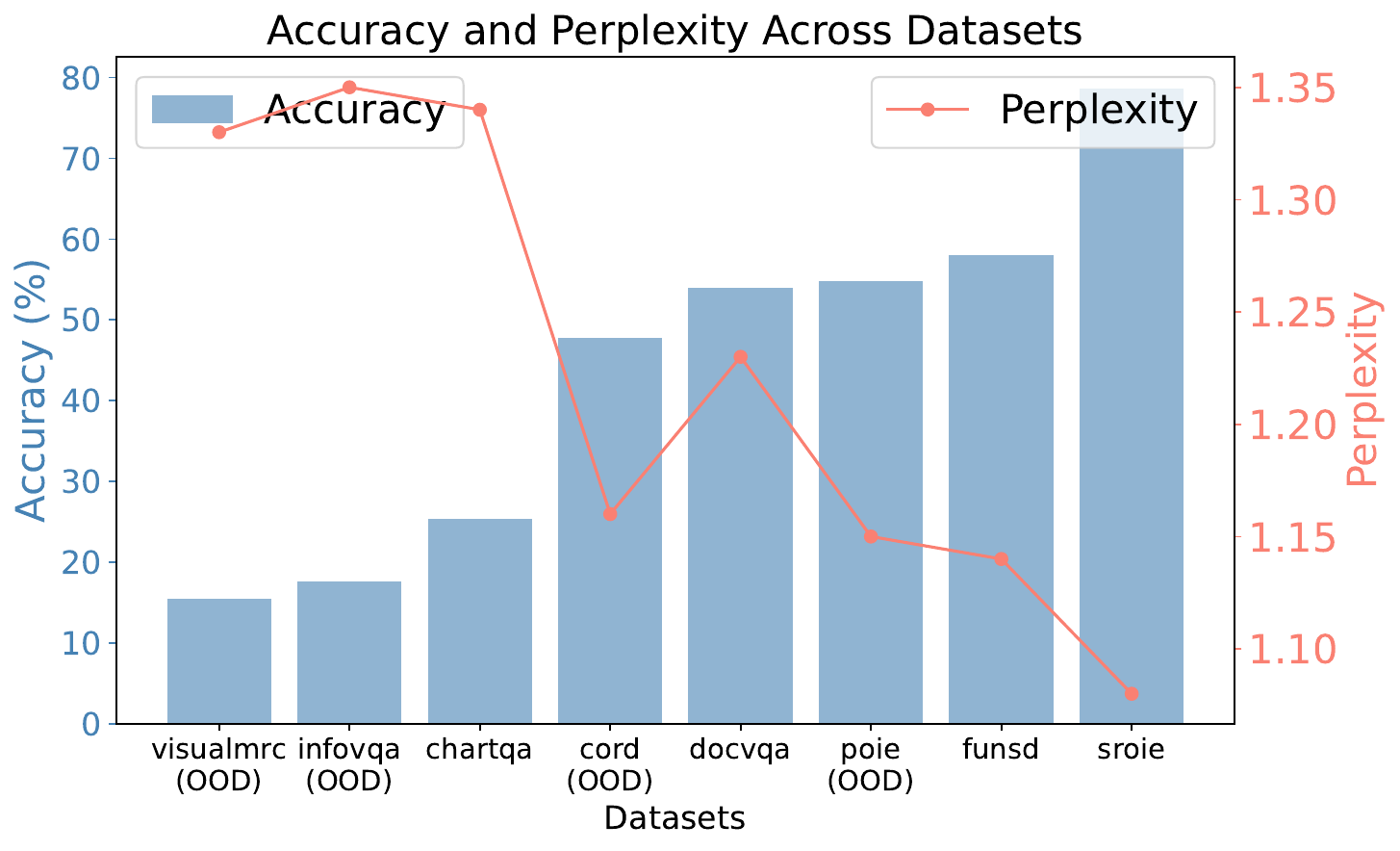}
  \caption{Mean accuracy vs. mean PPL for different datasets.}
  \label{fig:acc_vs_ppl}
\end{wrapfigure}

% \begin{figure*}[!t]
%     \centering
%     \subfigure[Qwen2.5-0.5B]{\includegraphics[width=.49\textwidth]{figure/0.5b/acc_ppl_0.5b.pdf}}
%     % \subfigure[Qwen2.5-7B]{\includegraphics[width=.49\textwidth]{figure/acc_ppl_7b.pdf}}
%     \caption{Mean accuracy vs. mean PPL for different datasets. OOD is out-of-domain test set.}
%     \label{fig:acc_vs_ppl}
% \end{figure*}

\begin{table}[t]
\centering
\scriptsize
\begin{tabular}{l|cccccccc|c}
\toprule
 \textbf{PPL} &  \textbf{DocVQA} & \textbf{ChartQA} & \textbf{InfoVQA} & \textbf{VisualMRC} & \textbf{FUNSD} & \textbf{POIE} & \textbf{CORD} & \textbf{SROIE} & \textbf{Average} \\
\midrule
Correct Cases & 1.15 & 1.17 & 1.25 & 1.28 & 1.09 & 1.09 & 1.14 & 1.05 & 1.15  \\
Incorrect Cases & 1.32 & 1.40 & 1.37 & 1.34 & 1.20 & 1.21 & 1.18 & 1.21 & 1.28 \\
\bottomrule
\end{tabular}
\caption{Mean PPL scores for correct and incorrect cases across datasets.}\label{tab:good_vs_bad_mean_ppl}
\end{table}

% \begin{figure*}[!t]
%     \centering
%     \subfigure[DocVQA]{\includegraphics[width=.24\textwidth]{figure/7b/docvqa_ppl.pdf}}
%     \subfigure[ChartQA]{\includegraphics[width=.24\textwidth]{figure/7b/chartqa_ppl.pdf}}
%     \subfigure[InfoVQA]{\includegraphics[width=.24\textwidth]{figure/7b/infovqa_ppl.pdf}}
%     \subfigure[VisualMRD]{\includegraphics[width=.24\textwidth]{figure/7b/visualmrc_ppl.pdf}}
%     \subfigure[FUNSD]{\includegraphics[width=.24\textwidth]{figure/7b/funsd_ppl.pdf}}
%     \subfigure[POIE]{\includegraphics[width=.24\textwidth]{figure/7b/poie_ppl.pdf}}
%     \subfigure[CORD]{\includegraphics[width=.24\textwidth]{figure/7b/cord_ppl.pdf}}
%     \subfigure[SROIE]{\includegraphics[width=.24\textwidth]{figure/7b/sroie_ppl.pdf}}
%     \caption{PPL distribution of Correct and Incorrect Predictions using Qwen2.5-7B. X-axis is PPL, Y-axis is frequency.}
%     \label{fig:good_bad_ppl_7b}
% \end{figure*}

 First, we analyze the relationship between accuracy and PPL scores at the dataset level, focusing on short-answer responses. Our analysis across multiple datasets reveals a strong inverse correlation, as illustrated in Figure~\ref{fig:acc_vs_ppl}: datasets with higher accuracy consistently exhibit lower average PPL. This pattern suggests that lower PPL (\textit{i.e.,} indicating higher confidence in model-generated answers) may be associated with greater accuracy. 
 
 Furthermore, we meticulously analyze examples with correct and incorrect predictions. By visualizing the PPL distributions and list the average PPL scores, we more intuitively verify this relationship (as shown in Figure~\ref{fig:good_bad_ppl_05b} and Table~\ref{tab:good_vs_bad_mean_ppl}). The PPL scores of correct answers are significantly lower than those of incorrect answers, showing a clear difference in distribution.

Given the negative correlation between PPL and prediction correctness, we propose an empirical method: using a PPL threshold to determine whether a prediction is confident or requires further reasoning. Specifically, if the PPL exceeds a predefined threshold (indicating low confidence), the model continues processing the question with long-form reasoning. In our experiments, we set the PPL threshold at the 75$^{th}$ percentile of the test set’s PPL distribution. Note that in real-world applications, the test set’s PPL scores are unavailable. Therefore, this experiment only serves as an exploratory validation of PPL-based confidence estimation. The results, shown in Table~\ref{tab:infer_ppl}, demonstrate that PPL thresholding achieves the best performance for the majority of datasets, highlighting the effectiveness of PPL as a confidence estimator for model outputs.
% Theoretically, if this approach leads to higher prediction accuracy, it would demonstrate the usefulness of PPL as a confidence estimator for model outputs.

The results of these exploratory experiments reveal the great potential of using PPL as an indicator for measuring the reliability of model outputs, demonstrating a close correlation between PPL and the correctness of short answers. Building on these exploratory findings, this paper will leverage them as a foundation to develop a dynamic reasoning decision-making framework using PPL. The goal is to create a model that can adaptively switch between short and long-form reasoning during inference. By avoiding redundant computations, this approach will significantly improve the model's reasoning efficiency and accuracy. The detailed methodology is presented in Section~\ref{sec:method}.

% Additionally, it will offer novel insights and methods to address the adaptability challenges faced by current reasoning-enhanced models. 

% Building on these exploratory findings, this paper will use them as a foundation to further explore how to utilize PPL to construct a dynamic reasoning decision-making framework. The goal is to obtain a model capable of adaptive switching between short and long-form reasnoning in inference, enabling accurate judgment of the required reasoning depth for each instance. This will avoid redundant computations, significantly enhance the model's reasoning efficiency and accuracy in complex tasks, and provide new ideas and methods for addressing the adaptability challenges of current reasoning-enhanced models. The detailed approach is presented in Section~\ref{sec:method}.

\begin{table}[!t]
\centering
\small
\begin{tabular}{l|cccccccccc}
\toprule
 \textbf{Method} &  \textbf{DocVQA} & \textbf{ChartQA} & \textbf{InfoVQA} & \textbf{VisualMRC} & \textbf{FUNSD} & \textbf{POIE} & \textbf{CORD} & \textbf{SROIE}  \\
\midrule
Short & 53.9& 25.3 & 17.6  & 15.5 & 58.0 & \textbf{54.8} &\textbf{47.7} & 78.6  \\
Long & 41.1 & 16.6 & 14.9 &  11.8 & 44.5 &30.6 & 34.8 & 67.5   \\
PPL & \textbf{54.2} & \textbf{26.1} & \textbf{18.4} & \textbf{15.8} & \textbf{59.3} & 53.7 & 38.7 & \textbf{80.9}  \\
\bottomrule
\end{tabular}
\caption{Performance of different inference methods across datasets.}\label{tab:infer_ppl}
\end{table}

\section{Methodology}~\label{sec:method}

\begin{figure*}  % 'r' for right side and width of the wrap figure
  \centering
  \includegraphics[width=0.9\columnwidth]{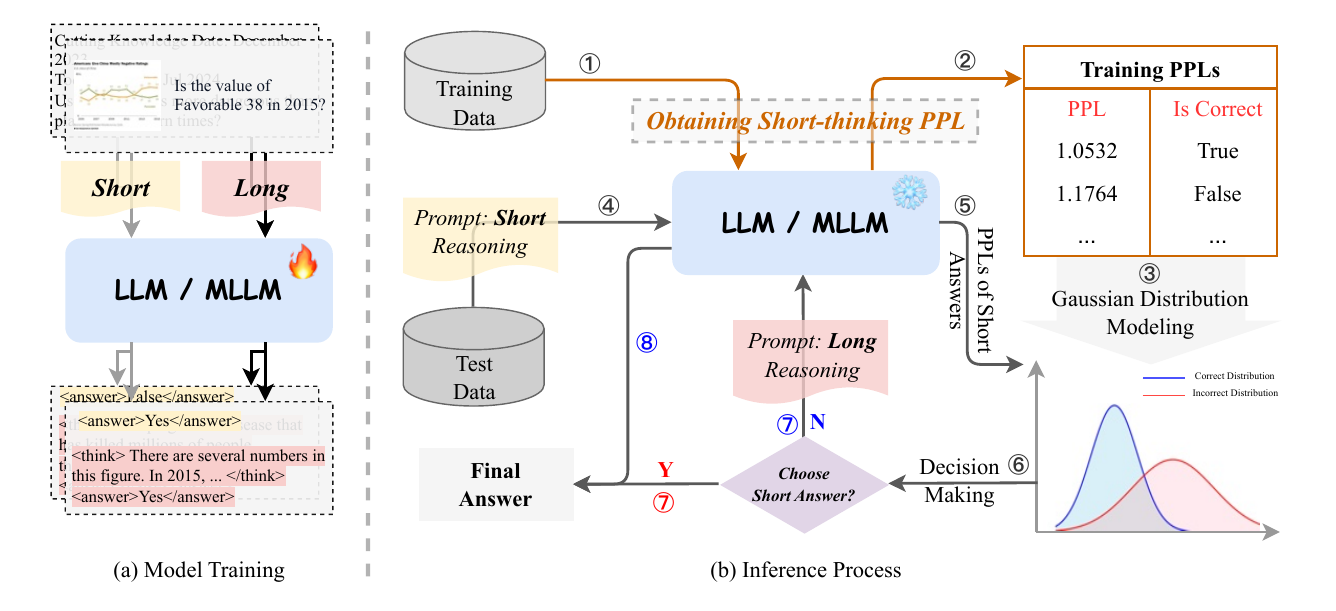}
  \caption{The training pipeline (a) and the inference process (b) of CAR.}
  \label{fig:framework}
\end{figure*}

As shown in Figure~\ref{fig:framework}(a), we first utilize both examples with short answers and examples with long-form reasoning answers for training LLMs or MLLMs. Subsequently, with the help of PPLs from training sets, we estimate the distributions of PPLs for correct and incorrect short answers, which are used to making decision. Specifically, if the estimated distribution determines that the short answer is correct, the proposed method outputs the correct answer directly. Otherwise, it performs the long-form reasoning. The inference process is shown in Figure~\ref{fig:framework}(b).

\subsection{Model Training}
% Let the training dataset be $\mathcal{D} = \{(x_i, y_{i}^{s}, y_{i}^{l})\}_{i = 1}^{N}$, where $x_i$ represents the $i$-th sample, $y_{i}^{s}$ is the corresponding short answer, and $y_{i}^{l}$ is the long-form reasoning answer.

With the training examples with each containing annotations for both short answers and long-form reasoning answers, we blend them to create a new dataset. To prompt the short answer, we use the instruction: \textit{``Please directly output the answer.''} For the long-form reasoning answer, we instead prompt: \textit{``Please output the reasoning process before outputting the answer.''} We then apply the standard instruction tuning procedure. The model takes a sequence of text $w_1, w_2, \ldots, w_T$ consisting of input and output text. The optimization objective is cross-entropy loss $\mathcal{L}$ which can be defined as follows: 

\begin{equation}
\mathcal{L}=-\frac{1}{T} \sum_{t=1}^T \log P\left(w_i \mid w_1, w_2, \ldots, w_{t-1}\right)
\end{equation}

\noindent where $P\left(w_i \mid w_1, w_2, \ldots, w_{t-1}\right)$ represents the probability of $t^{th}$ token $w_{t}$ given the sequence of preceding tokens $w_1, w_2, \ldots, w_{t-1}$. We only compute the loss for the response tokens.

% We use the Qwen 2.5 0.5b model as the base model and fine-tune it on this dataset to obtain a model $M$ that can generate both short and long-form reasoning responses. The training process aims to minimize a loss function $\mathcal{L}$, which can be a combination of cross-entropy losses for short and long-form reasoning answers:

% \begin{equation}
% \mathcal{L}=\alpha\mathcal{L}_{s}+(1 - \alpha)\mathcal{L}_{l}
% \end{equation}

% where $\alpha\in[0,1]$ is a hyperparameter that balances the contributions of short response loss $\mathcal{L}_{s}$ and long-form reasoning loss $\mathcal{L}_{l}$. $\mathcal{L}_{s}$ measures the difference between the model's short output and $y_{i}^{s}$, and $\mathcal{L}_{l}$ measures the difference between the long-form reasoning output and $y_{i}^{l}$.

\subsection{Obtaining Short Answer PPL}
After the training of model $M$ is completed, we perform short answer inference on all examples in the training dataset $\mathcal{D}$. For each example $x_i$, the model generates a short answer $\hat{y}_{i}^{s}$, and we calculate its PPL value $PPL_i$. The PPL of a sequence of tokens ($w_1, w_2,\cdots, w_T$) is defined as:

% PPL is one of the most widely used metrics for evaluating language models, defined as the exponential average of the negative log-likelihood of a sequence:

\begin{equation}
PPL = \exp\left(-\frac{1}{T}\sum_{t = 1}^{T}\log p(w_t|w_1,\cdots,w_{t - 1})\right)
\end{equation}\label{eq:ppl}

where $p(w_t|w_1,\cdots,w_{t - 1})$ is the probability of the $t$-th token given the previous $t - 1$ tokens predicted by the model. 

% Intuitively, for a given language model, a lower PPL score on a corpus of text suggests that the model is better at capturing the underlying patterns in the data. 

\subsection{Gaussian Distribution Modeling}
Let $C$ be a binary variable indicating whether the short answer is correct or not, where $C = 1$ represents a correct answer and $C = 0$ represents an incorrect answer. We assume that the distribution of PPL scores for correct and incorrect short answers follows a Gaussian distribution.

For correct answers $C = 1$, the PPL scores follow a Gaussian distribution $\mathcal{N}(\mu_1,\sigma_1^{2})$, and for incorrect answers $C = 0$, the PPL scores follow a Gaussian distribution $\mathcal{N}(\mu_0,\sigma_0^{2})$. The probability density functions are given by:

\begin{equation}
f_1(PPL)=\frac{1}{\sqrt{2\pi}\sigma_1}\exp\left(-\frac{(PPL-\mu_1)^{2}}{2\sigma_1^{2}}\right)
\end{equation}

\begin{equation}
f_0(PPL)=\frac{1}{\sqrt{2\pi}\sigma_0}\exp\left(-\frac{(PPL-\mu_0)^{2}}{2\sigma_0^{2}}\right)
\end{equation}

We estimate the parameters $\mu_1,\sigma_1^{2},\mu_0,\sigma_0^{2}$ from the training data. Let $n_1$ be the number of correct short answers and $n_0$ be the number of incorrect short answers in the training set. The parameter estimates are:

\begin{equation}
\hat{\mu}_1=\frac{1}{n_1}\sum_{i:C_i = 1}PPL_i,\quad\hat{\sigma}_1^{2}=\frac{1}{n_1}\sum_{i:C_i = 1}(PPL_i-\hat{\mu}_1)^{2}
\end{equation}

\begin{equation}
\hat{\mu}_0=\frac{1}{n_0}\sum_{i:C_i = 0}PPL_i,\quad\hat{\sigma}_0^{2}=\frac{1}{n_0}\sum_{i:C_i = 0}(PPL_i-\hat{\mu}_0)^{2}
\end{equation}

\subsection{Inference Process}
When the trained model $M$ is used for inference on a new input example $x$, the following steps are carried out: \textbf{1) Short Answer inference}: The model $M$ first generates a short answer $\hat{y}^{s}$ for the input $x$, and we calculate its PPL score, denoted as $PPL_{new}$. \textbf{2) Probability calculation}: We substitute $PPL_{new}$ into the probability density functions $f_1$ and $f_0$ to obtain the probabilities $P(C = 1|PPL_{new})$ and $P(C = 0|PPL_{new})$. According to Bayes' theorem:
\begin{equation}
P(C = 1|PPL_{new})=\frac{f_1(PPL_{new})\cdot P(C = 1)}{f_1(PPL_{new})\cdot P(C = 1)+f_0(PPL_{new})\cdot P(C = 0)}
\end{equation}
\begin{equation}
P(C = 0|PPL_{new})=\frac{f_0(PPL_{new})\cdot P(C = 0)}{f_1(PPL_{new})\cdot P(C = 1)+f_0(PPL_{new})\cdot P(C = 0)}
\end{equation}
where $P(C = 1)=\frac{n_1}{n_1 + n_0}$ and $P(C = 0)=\frac{n_0}{n_1 + n_0}$ are the prior probabilities of correct and incorrect answers in the training set. To balance the correctly-answered and incorrectly-answered examples, we select the same number of correctly-answered examples as the incorrectly answered ones for the aforementioned modeling, \textit{i.e.}, $n_0=n_1$. \textbf{3) Decision-making}: If $P(C = 1|PPL_{new})>P(C = 0|PPL_{new})$, we consider the short answer $\hat{y}^{s}$ to be accurate enough and output it as the final answer. Otherwise, we use the long-form reasoning prompt to guide the model $M$ to perform long-form reasoning and obtain a more accurate answer $\hat{y}^{l}$, which is then output as the final result.

\section{Experiments}~\label{sec:main_exp}

% In this section, we showcase the effectiveness of CAR in terms of prediction quality and token reduction through experiments.

\subsection{Setup}\label{subsec:setup}

\paragraph{Datasets} To demonstrate the utility of our proposed method, we conduct experiments on three multimodality datasets: \textbf{DocVQA}~\cite{mathew2021docvqa}, \textbf{ChartQA}~\cite{masry-etal-2022-chartqa}, and \textbf{FUNSD}~\cite{jaume2019funsd}. Unlike the pilot study in Section~\ref{sec:pilot_study}, here we input the image modality and evaluate the performance using MLLMs. Since these datasets lack reasoning process annotations, we reuse the reasoning process data obtained in the pilot study. Furthermore, we evaluate CAR on text datasets, selecting three widely used reasnong datasets: the mathematical reasoning datasets \textbf{GSM8K}~\cite{cobbe2021training} and \textbf{MathQA}~\cite{amini-etal-2019-mathqa}, as well as the commonsense reasoning dataset \textbf{StrategyQA}~\cite{geva2021did}.

\paragraph{Training and inference setup}

We utilize Qwen2-VL-7B-Instruct~\cite{wang2024qwen2} as our multimodal language model, and employ Qwen2.5-7B-Instruct~\cite{yang2024qwen2} and Llama3.1-8B-Instruct~\cite{grattafiori2024llama} as large language models, termed as \textbf{CAR$_{Qwen2VL}$}, \textbf{CAR$_{Qwen2.5}$} and \textbf{CAR$_{Llama3.1}$} respectively. We train all models for 3 epochs, using a batch size of 32 and a learning rate of 1e-6, with the AdamW optimizer~\cite{loshchilov2017decoupled} and a cosine scheduler~\cite{loshchilov2016sgdr}. The maximum input and output
sequence lengths are set to 4096 and 1024, respectively. Training is conducted on 8 NVIDIA A100.

To eliminate the impact of randomness, no sampling methods are employed during testing for any of the models. Instead, beam search with a beam size of 1 is used for generation across all models. Additionally, the maximum number of generated tokens is set to 1024, while the maximum number of input tokens is set to 4096.

\begin{table*}[!t]
\centering
\resizebox{0.85\textwidth}{!}{%
\small
\begin{tabular}{l|cccccc|cc}
\toprule
~ & \multicolumn{2}{c}{\textbf{DocVQA}} & \multicolumn{2}{c}{\textbf{ChartQA}} & \multicolumn{2}{c}{\textbf{FUNSD}} & \multicolumn{2}{c}{\textbf{Average}} \\  \midrule
\textbf{Model} & Acc & \#Token & Acc & \#Token & Acc & \#Token & Acc & \#Token \\\midrule
\textbf{Baselines}               &                &                &                        &                 &                &       &       &      \\\midrule
Qwen2-VL\textsubscript{Short}~\cite{wang2024qwen2} & 87.4 & 10.7 & 66.6 & 9.0 & 71.5& 16.7 & 75.1 & 12.1 \\
Qwen2-VL\textsubscript{Long}~\cite{wang2024qwen2} & 84.6 & 404.1 & 65.2 & 978.1 & 67.6 &  346.8 &72.4& 576.3 \\\midrule
% \rowcolor{gray!20} CAR\textsubscript{UpperBound} & 92.4 & 110.7 & 75.2 & 182.4 & 76.0 & 171.7 & 81.2 & 154.9 \\
\textbf{Ours}               &                &                &                        &                 &                &       &       &      \\\midrule
CAR\textsubscript{Short} & 89.0 & 10.5 & 67.1 & 9.0 & 71.5 & 16.8 & 75.8 & 12.1 \\
CAR\textsubscript{Long} & 83.9 & 426.6 & 65.6 & 394.2 & 67.4  & 469.8 & 72.3 & 430.2 \\
CAR\textsubscript{Qwen2VL} & \textbf{90.1} & 88.9 & \textbf{69.9} & 80.0 &\textbf{73.6} & 91.8 & \textbf{77.9} & 86.9 \\
% Qwen2.5-0.5B\textsubscript{Short}                        &        \textbf{56.83}        &         \textbf{31.92}         &         \textbf{58.02}        &   \textbf{48.92} \\
% Qwen2.5-0.5B\textsubscript{Long}                       &           44.01        &   23.52      &            44.53   &  37.35
\bottomrule
\end{tabular}%
}
\caption{Comparison of accuracy and token numbers in multimodality datasets.}
\label{tab:main_result_mllm}
\end{table*}

\paragraph{Baselines}

The baselines used in our experiments include: 1) \textbf{Short answer and long-form reasoning baselines:} This work primarily investigates whether the CAR method can outperform inference based fully on short-answer or long-form reasoning. To evaluate this, we fine-tune Qwen2-VL-7B-Instruct with the same hyperparameters on both short-answer and long-form response annotations, resulting in two variants: \textbf{Qwen2-VL$_{Short}$}, \textbf{Qwen2-VL$_{long}$}. Since CAR can perform both short-answer and long-form reasoning, we also report its results under these two inference modes, denoted as \textbf{CAR$_{Short}$} and \textbf{CAR$_{Long}$} respectively. 2) \textbf{SOTA reasoning token reduction baselines:} Among reasoning methods, those most closely related to CAR are reasoning token reduction methods in LLM~\cite{sui2025stop}. As strong baselines, we adopt two recent state-of-the-art LLM-based techniques: Token-Budget-Aware Reasoning (TALE)~\cite{han2024token} and Chain-of-Draft (COD)~\cite{xu2025chain}. We replicate these methods using Qwen2.5-7B-Instruct and Llama3.1-8B-Instruct, resulting in four variants: \textbf{TALE$_{Qwen2.5}$}, \textbf{TALE$_{Llama3.1}$}, \textbf{COD$_{Qwen2.5}$} and \textbf{COD$_{Llama3.1}$}. To establish direct comparisons, we also fine-tune both Qwen2.5-7B-Instruct and Llama3.1-8b-Instruct using identical hyperparameters on short-answer and long-form response annotations, resulting in four variants: \textbf{Qwen2.5$_{Short}$}, \textbf{Qwen2.5$_{long}$}, \textbf{Llama3.1$_{Short}$} and \textbf{Llama3.1$_{long}$}. We provide detailed implementation configurations and hyperparameters for these baselines in Appendix~\ref{app:exp_setup}.

\paragraph{Evaluation metric} Our evaluation focuses on two key dimensions: prediction quality and inference efficiency. For prediction quality, we adopt task-specific metrics. On VQA and KIE datasets, we follow~\cite{feng2023unidoc,luo2024layoutllm,feng2023docpedia} by using accuracy, where a response is considered correct if it contains the ground truth. For mathematical and commonsense reasoning tasks, as in~\cite{touvron2023llama,wang2024qwen2,qwen7b}, we require exact matches the ground truth for correctness. To measure inference efficiency, we record the number of generated tokens as our metric.

\begin{table*}[t]
\centering
\resizebox{0.85\textwidth}{!}{%
\small
\begin{tabular}{l|cccccc|cc}
\toprule
~ & \multicolumn{2}{c}{\textbf{GSM8K}} & \multicolumn{2}{c}{\textbf{StrategyQA}} & \multicolumn{2}{c}{\textbf{MathQA}} & \multicolumn{2}{c}{\textbf{Average}} \\  \midrule
\textbf{Model} & Acc & \#Token & Acc & \#Token & Acc & \#Token & Acc & \#Token \\
\midrule
\textbf{Baselines} &  &  &  & &  &  &  &  \\
\midrule
% \rowcolor{gray!20} Upper Bound & 92.4 & - & 75.2 & - & 76.0 & - & - & - \\
Qwen2.5\textsubscript{Short}~\cite{yang2024qwen2} & 24.2 & 9.2 & 78.2 & 8.0 & 65.0 & 8.0 & 55.8 & 8.4 \\
Qwen2.5\textsubscript{Long}~\cite{yang2024qwen2} & 76.4 & 138.4 & 76.8 & 71.7 & 72.0 & 168.4 & 75.0 & 126.1 \\
\midrule
\textbf{SOTA} &  &  &  & &  &  &  &  \\
\midrule
% \rowcolor{gray!20} Upper Bound & 92.4 & - & 75.2 & - & 76.0 & - & - & - \\
TALE\textsubscript{Qwen2.5}~\cite{han2024token} & \textbf{87.6} & 124.0 & 80.6 & 51.7 & 50.4 & 321.7 & 72.8 & 165.8 \\
COD\textsubscript{Qwen2.5}~\cite{xu2025chain} & 78.9 & 93.4 & 76.8 & 31.1 & 67.1 & 139.6 & 74.2 & 88.1 \\
\midrule
\textbf{Ours} &  &  &  & &  &  &  &  \\
\midrule
% \rowcolor{gray!20}
% CAR\textsubscript{Qwen2.5-UpperBound} & 80.1 & 115.0 & 89.0 & 18.1 & 87.4 & 47.4 & 85.5 & 60.2 \\
CAR\textsubscript{Qwen2.5-Short} & 23.6 & 9.4 & 78.3 & 8.0 & 65.4 & 8.0 & 55.8 & 8.5 \\
CAR\textsubscript{Qwen2.5-Long} & 76.8 & 139.4 & 79.0 & 74.1 & 80.6 & 130.1  & 78.8 & 114.5 \\
CAR\textsubscript{Qwen2.5} & 78.8 & 127.8 & \textbf{80.7} & 21.8 &\textbf{83.8} & 58.1 & \textbf{81.1} & 69.2 \\
% Qwen2.5-0.5B\textsubscript{Short}                        &        \textbf{56.83}        &         \textbf{31.92}         &         \textbf{58.02}        &   \textbf{48.92} \\
% Qwen2.5-0.5B\textsubscript{Long}                       &           44.01        &   23.52      &            44.53   &  37.35
\bottomrule
\end{tabular}%
}
\caption{Comparison of accuracy and token numbers in reasoning datasets using Qwen2.5.}
\label{tab:exp_qwen}
\end{table*}

\begin{table*}[t]
\centering
\resizebox{0.85\textwidth}{!}{%
\small
\begin{tabular}{l|cccccc|cc}
\toprule
~ & \multicolumn{2}{c}{\textbf{GSM8K}} & \multicolumn{2}{c}{\textbf{StrategyQA}} & \multicolumn{2}{c}{\textbf{MathQA}} & \multicolumn{2}{c}{\textbf{Average}} \\  \midrule
\textbf{Model} & Acc & \#Token & Acc & \#Token & Acc & \#Token & Acc & \#Token \\
\midrule
\textbf{Baselines} &  &  &  & &  &  &  &  \\
\midrule
% \rowcolor{gray!20} Upper Bound & 92.4 & - & 75.2 & - & 76.0 & - & - & - \\
Llama3.1\textsubscript{Short}~\cite{grattafiori2024llama} & 23.9 & 9.1 & 75.5 & 9.0 & 54.4& 9.0 & 33.1 & 9.0 \\
Llama3.1\textsubscript{Long}~\cite{grattafiori2024llama} & 72.9 & 118.9 & 78.6 & 175.9 & 60.9 & 220.8 & 70.8 & 171.8 \\
\midrule
\textbf{SOTA} &  &  &  & &  &  &  &  \\
\midrule
% \rowcolor{gray!20} Upper Bound & 92.4 & - & 75.2 & - & 76.0 & - & - & - \\
TALE\textsubscript{Llama3.1}~\cite{han2024token} & \textbf{80.6} & 183.2 & 76.2 & 89.1 & 48.3 & 639.6 & 68.3 & 303.9 \\
COD\textsubscript{Llama3.1}~\cite{xu2025chain} & 73.7 & 117.7 & 75.5 & 73.7 & 59.1 & 268.7 & 69.4 & 153.3 \\
\midrule
% \rowcolor{gray!20}
% CAR\textsubscript{Llama3.1-UpperBound} & 77.4 & 99.4 & 88.3 & 40.4 & 81.9 & 102.7 & 82.5 & 80.8 \\
\textbf{Ours} &  &  &  & &  &  &  &  \\
\midrule
CAR\textsubscript{Llama3.1-Short} & 21.3 & 9.1 & 75.9 & 9.0 & 57.3 & 9.0 & 51.5 & 9.0 \\
CAR\textsubscript{Llama3.1-Long} & 73.5 & 116.2 & 81.4 & 90.9 & \textbf{73.1} & 227.3  & \textbf{76.0} & 144.8 \\
CAR\textsubscript{Llama3.1} & 71.6 & 108.9 & \textbf{82.8} & 21.4 & 70.2 & 149.9 & 74.9 & 93.4 \\
% Qwen2.5-0.5B\textsubscript{Short}                        &        \textbf{56.83}        &         \textbf{31.92}         &         \textbf{58.02}        &   \textbf{48.92} \\
% Qwen2.5-0.5B\textsubscript{Long}                       &           44.01        &   23.52      &            44.53   &  37.35
\bottomrule
\end{tabular}%
}
\caption{Comparison of accuracy and token numbers in reasoning datasets using Llama3.1.}
\label{tab:exp_llama}
\end{table*}

\subsection{Results on Multimodal Datasets}

Table~\ref{tab:main_result_mllm} demonstrates the performance on multimodal datasets. First, the superior performance of CAR$_{Qwen2VL}$ over CAR$_{Short}$ and CAR$_{Long}$ demonstrates the effectiveness of using PPL as an indicator for routing reasoning paths. Furthermore, the CAR$_{Qwen2VL}$ achieves the highest average accuracy 77.9\%, outperforming baseline models Qwen2VL$_{Short}$ and Qwen2VL$_{Long}$ by 2.8\% and 5.5\% improvement respectively. Notably, our method maintains efficient token usage (86.9 tokens on average), requiring only 15\% of the tokens used by Qwen2VL$_{Long}$. These results showcase the utility of CAR in multimodal scenarios.

\subsection{Results on Text Datasets}

Table~\ref{tab:exp_qwen} and~\ref{tab:exp_llama} show the performance on text-based reasoning tasks. CAR demonstrates robust performance, achieving an average accuracy of 81.1\% with Qwen2.5-7B and 74.9\% with Llama3.1-8B, outperforming both short-answer baselines (55.8\% and 51.5\%) and exhaustive long-form reasoning (75.0\% and 70.8\%). Notably, CAR reduces token usage by 45.1\% (Qwen2.5) and 45.6\% (Llama3.1) compared to pure long-form reasoning. In Qwen2.5, CAR$_{Qwen2.5}$ consistently outperforms CAR$_{Qwen2.5-Short}$ and CAR$_{Qwen2.5-Long}$, again demonstrating the effectiveness of using PPL as a routing indicator.

Furthermore, CAR consistently outperforms state-of-the-art token reduction methods like TALE and COD. Specifically, with Qwen2.5, CAR achieves an average accuracy 8.3\% higher than TALE and 6.9\% higher than COD, while maintaining the lowest token count (\textit{i.e.,} 69.2 tokens). Similarly, with Llama3.1, CAR surpasses TALE and COD by 6.6\% and 5.5\% in average accuracy, respectively, while still generating the fewest tokens.

Notably, CAR’s adaptive routing proves particularly effective in MathQA datasets (\textit{i.e.,} 70.2\% vs. 59.1\% of COD with Llama3.1, 83.8\% vs. 67.1\% of COD with Qwen2.5), where it eliminates unnecessary reasoning steps. These results underscore CAR’s utility across diverse reasoning paradigms.

\subsection{Discussion}

Our experiments demonstrate that CAR achieves significant improvements on multimodal VQA/KIE datasets, while showing more modest gains on complex reasoning tasks like GSM8K and MathQA. This discrepancy aligns with our hypothesis: tasks like GSM8K inherently require CoT reasoning for most questions, leaving limited room for improvement when switching to short answers. In fact, this observation further validates CAR’s reliability—it correctly identifies that exhaustive reasoning is necessary for such tasks, avoiding unnecessary shortcuts that would harm accuracy.

Notably, while CAR slightly underperforms TALE on GSM8K (78.8\% vs. 87.6\%), the two methods are complementary rather than contradictory. TALE optimizes reasoning efficiency by generating shorter reasoning paths via prompt engineering and SFT, whereas CAR dynamically routes between short answers and full reasoning based on confidence. This suggests a promising direction: combining CAR’s adaptive routing with TALE’s concise reasoning generation could yield even higher accuracy and efficiency. More experiments in Appendix~\ref{app:combined_method} show the performance of such hybrid approach.

These insights highlight the university of CAR that it not only excels in tasks where reasoning is selectively beneficial (\textit{e.g.,} VQA and KIE) but also effective in tasks requiring reasoning such as MathQA and StrategyQA. Future work could explore tighter integration of confidence-aware routing with other reasoning token reduction methods.

\section{Limitations}\label{sec:limit}

While CAR effectively balances accuracy and efficiency, it has several limitations: (1) Performance gains are more modest on complex reasoning tasks (\textit{e.g.,} MathQA, GSM8K) compared to fact-based VQA/KIE datasets, suggesting room for improvement in tasks inherently require reasoning; (2) The adaptive routing process introduces minor computational overhead compared to pure short-answer inference. 

\section{Conclusion}

We propose \textbf{C}ertainty-based  \textbf{A}daptive  \textbf{R}easoning (CAR), a framework that dynamically switches between short answers and long-form reasoning based on model confidence. Experiments across diverse benchmarks demonstrate CAR’s effectiveness in balancing efficiency and performance. Our work challenges the necessity of always using long-form reasoning, offering a more adaptive and efficient approach for LLMs/MLLM reasoning. 

\bibliographystyle{unsrtnat}
\bibliography{neurips_2025}

\newpage
%%%%%%%%%%%%%%%%%%%%%%%%%%%%%%%%%%%%%%%%%%%%%%%%%%%%%%%%%%%%

\appendix

\section{Preliminary Experiment Setup}\label{app:intro_exp}

\subsection{Data Creation}

For the polit experiment, we conduct experiments on the same datasets as mentioned in Section~\ref{sec:pilot_study}, \textit{i.e.,} VQA datasets \textbf{DocVQA}~\citep{mathew2021docvqa}, \textbf{ChartQA}~\citep{masry-etal-2022-chartqa} and KIE datasets \textbf{SROIE}~\citep{huang2019icdar2019}, \textbf{FUNSD}~\citep{jaume2019funsd}. Since these datasets are multimodal but the pilot study focuses on LLM-based reasoning, we structure the input using the OCR text and bounding box provided by respective dataset. For example: ``Given document <document>Saint[466,77,554,107]\textbackslash nLouis[561,77,657,107]\textbackslash nHeart[664,77,760,106] ...''. However, these datasets do not include reasoning steps. To address this, we employ DeepSeek-R1 to generate plausible reasoning processes by providing the question and final answer. The statistics is shown in Table~\ref{tab:data_statis}.

\begin{table}[t]
\centering
\small
\begin{tabular}{l|cccc|c}
\toprule
 \textbf{PPL} &  \textbf{DocVQA} & \textbf{ChartQA}  & \textbf{FUNSD}  & \textbf{SROIE} & \textbf{Average} \\
\midrule
Short Answer & 11.6 & 9.1  & 17.2 & 21.7 & 14.9  \\
Long-form Thinking & 472.8 & 1156.6  & 612.6 & 330.2 & 643.1 \\
\bottomrule
\end{tabular}
\caption{Token count of short answer and long-form thinking annotations using Qwen2.5 tokenizer.}\label{tab:data_statis}
\end{table}

\subsection{Setup}

We fine-tune Qwen2.5-0.5B and Qwen2.5-7B on DocVQA, ChartQA, FUNSD, and SROIE, evaluating performance on these in-domain datasets as well as on out-of-domain (OOD) test sets (POIE, InfoVQA, VisualMRC, and CORD). For each example, we train the models to generate two response types: concise short answers when prompted with \textit{``Please directly output the answer''} and long-form reasoning followed by answers when prompted with \textit{``Please output the reasoning process before outputting the answer''}. We train all models for 3 epochs, using a batch size of 32 and a learning rate of 1e-6, with the AdamW optimizer~\cite{loshchilov2017decoupled} and a cosine scheduler~\cite{loshchilov2016sgdr}. The maximum input and output
sequence lengths are set to 4096 and 1024, respectively. Training is conducted on 8 NVIDIA A100.

\section{Visualization of Reasoning Cases}\label{app:intro_vis}

\begin{figure*}[ht]
	\centering
	\includegraphics[width=1.0\textwidth]{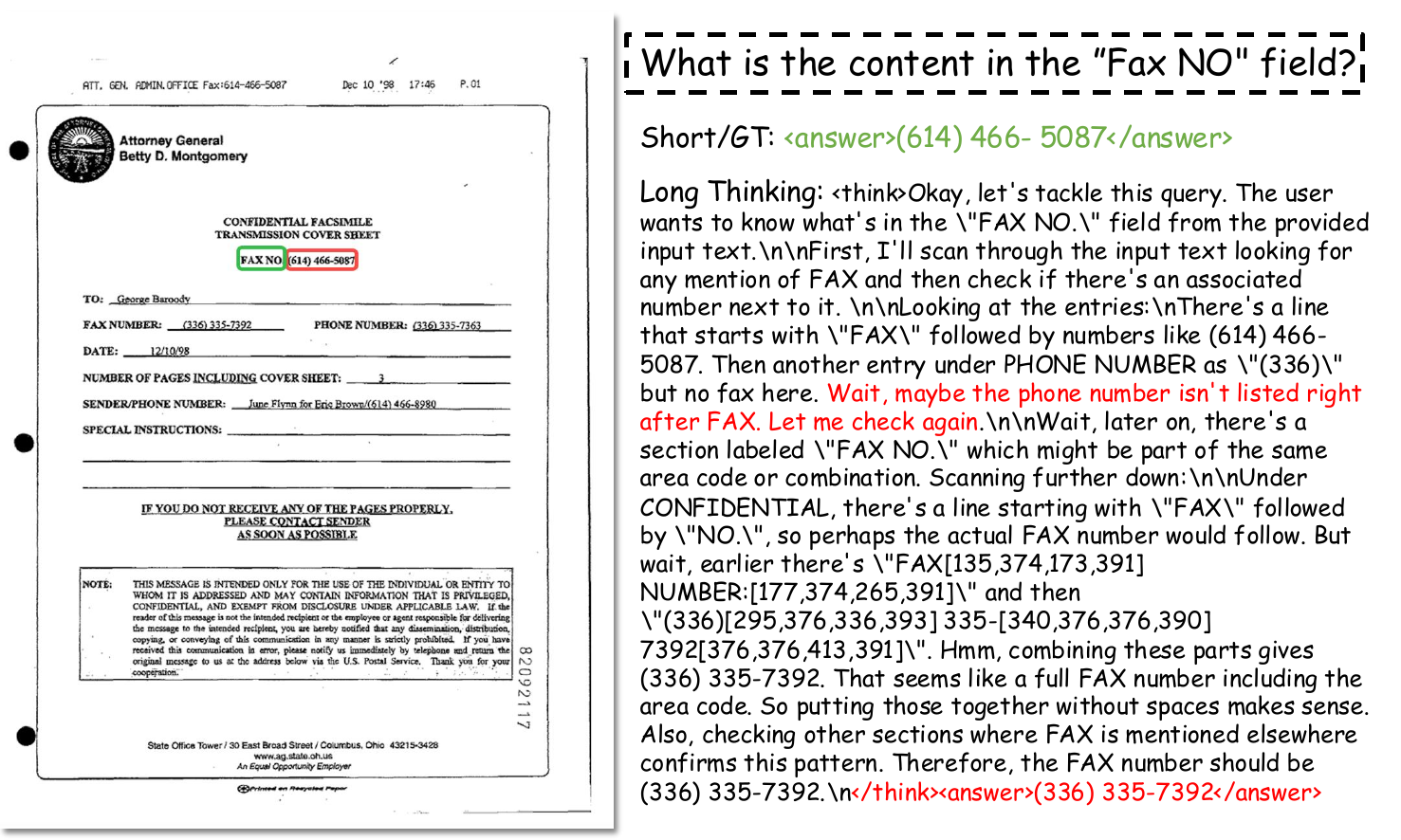}
	\caption{A failure case of reasoning model.}
	\label{fig:failure_case_funsd1}
\end{figure*}

\begin{figure*}[ht]
	\centering
	\includegraphics[width=1.0\textwidth]{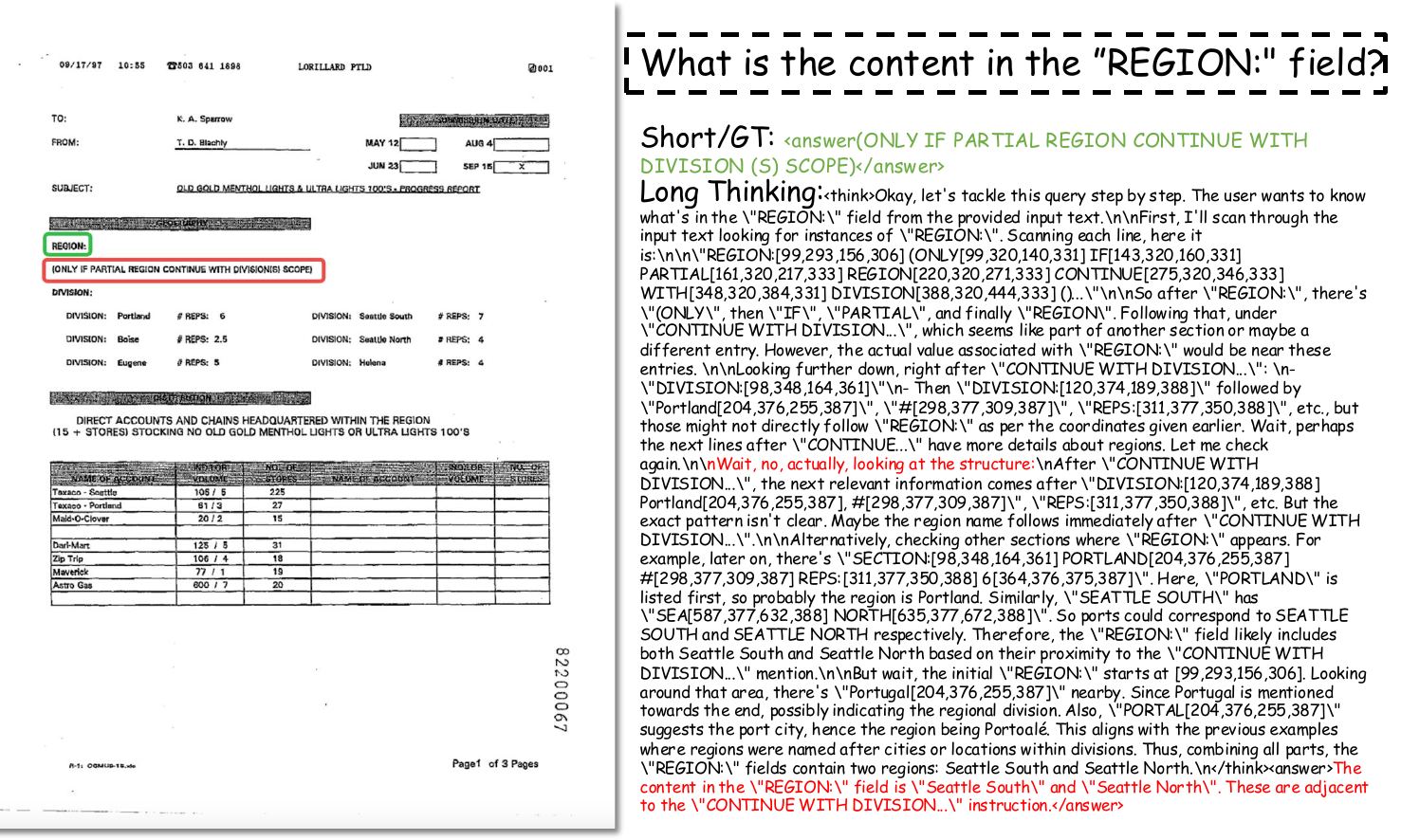}
	\caption{A failure case of reasoning model.}
	\label{fig:failure_case_funsd2}
\end{figure*}

The visualizations in Figure~\ref{fig:failure_case_funsd1} and~\ref{fig:failure_case_funsd2} compare the prediction results of the Qwen2.5-0.5B reasoning model and the short answer model. In the left image, the green and red rectangles highlight the key and the ground truth extracted value, respectively.

For these two questions, the short answer model easily retrieves the correct answer, matching the ground truth. In contrast, the long-form reasoning model initially identifies the right answer during early inference (as seen in its initial output). However, after further reflection—indicated by the red text—it ultimately arrives at an incorrect conclusion.

This suggests that while the reasoning model engages in deeper analysis, its additional processing can sometimes lead it away from the correct answer, unlike the more straightforward short answer model.

\section{Baseline Configurations}\label{app:exp_setup}

For all baselines, we follow the same training hyperparameters, that is, 3 epochs, using a batch size of 32 and a learning rate of 1e-6, with the AdamW optimizer~\cite{loshchilov2017decoupled} and a cosine scheduler~\cite{loshchilov2016sgdr}. The maximum input and output sequence lengths are set to 4096 and 1024, respectively.

For TALE~\cite{han2024token} method, we replicate the SFT version to ensure a fair comparison.

\section{Pilot Study Results on Qwen2.5-7B}\label{app:7b_result}

This section reports the same experiments as in Section~\ref{sec:pilot_study}, but using the Qwen2.5-7B model. The overall results align with those in the pilot study.

\begin{wrapfigure}{!t}{0.5\textwidth}  % 'r' for right side and width of the wrap figure
  \centering
  \includegraphics[width=.5\columnwidth]{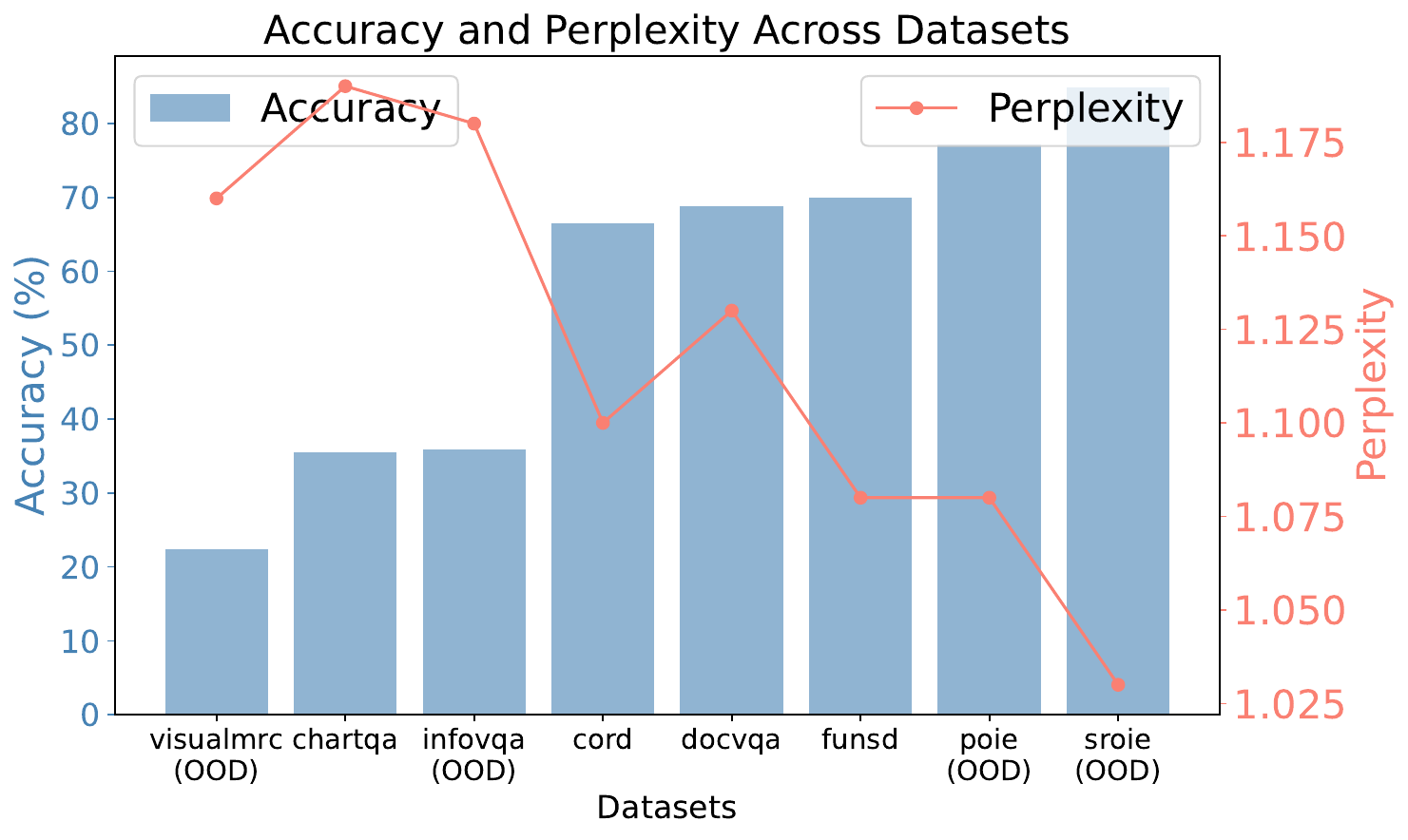}
  \caption{Mean accuracy vs. mean PPL for different datasets. OOD is out-of-domain test set.}
  \label{fig:acc_vs_ppl_7b}
\end{wrapfigure}

\begin{figure*}[t]
    \centering
    \subfigure[DocVQA]{\includegraphics[width=.24\textwidth]{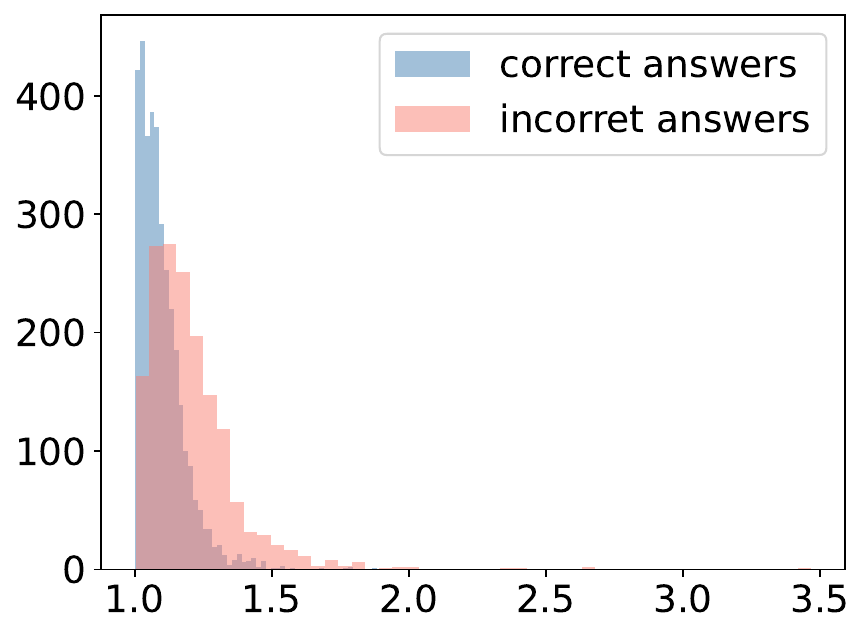}}
    \subfigure[ChartQA]{\includegraphics[width=.24\textwidth]{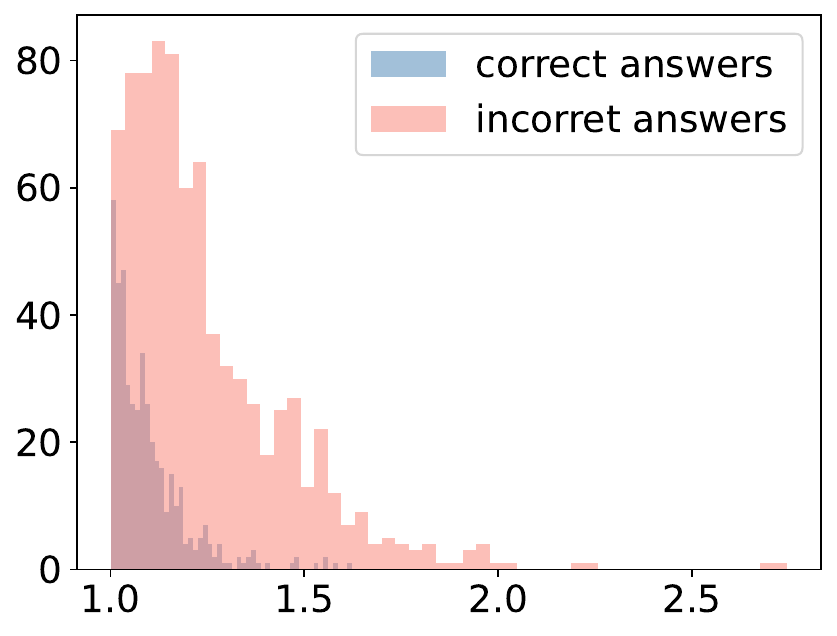}}
    \subfigure[InfoVQA(OOD)]{\includegraphics[width=.24\textwidth]{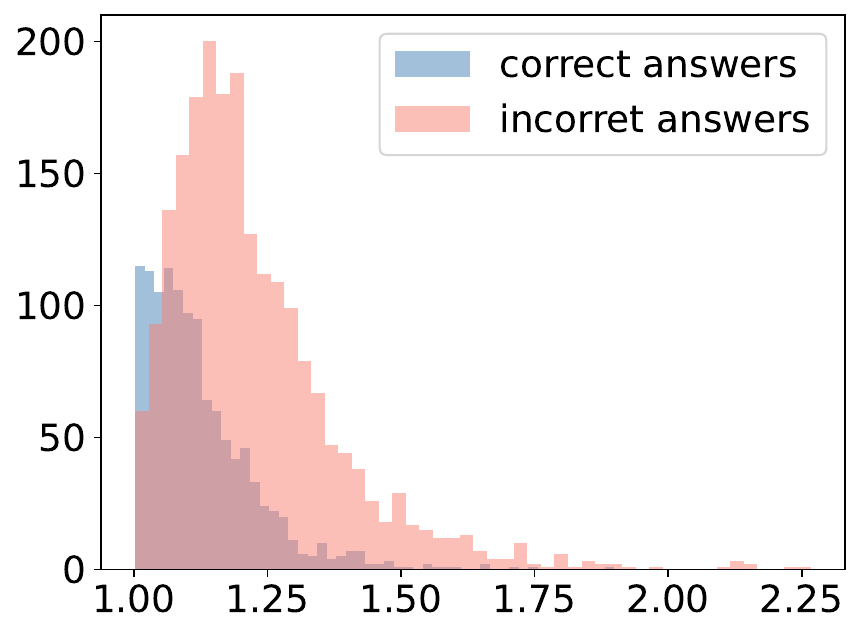}}
    \subfigure[VisualMRC(OOD)]{\includegraphics[width=.24\textwidth]{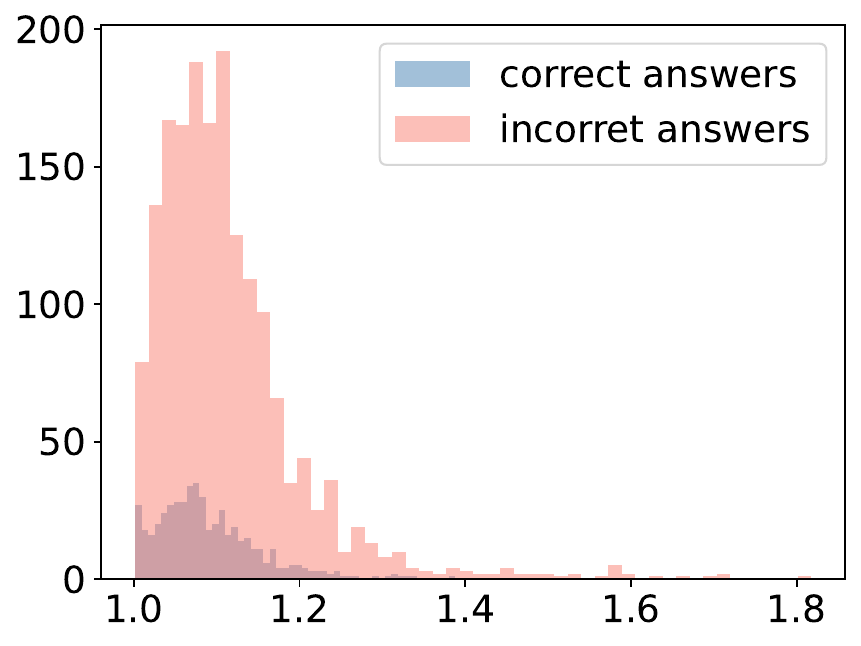}}
    \subfigure[FUNSD]{\includegraphics[width=.24\textwidth]{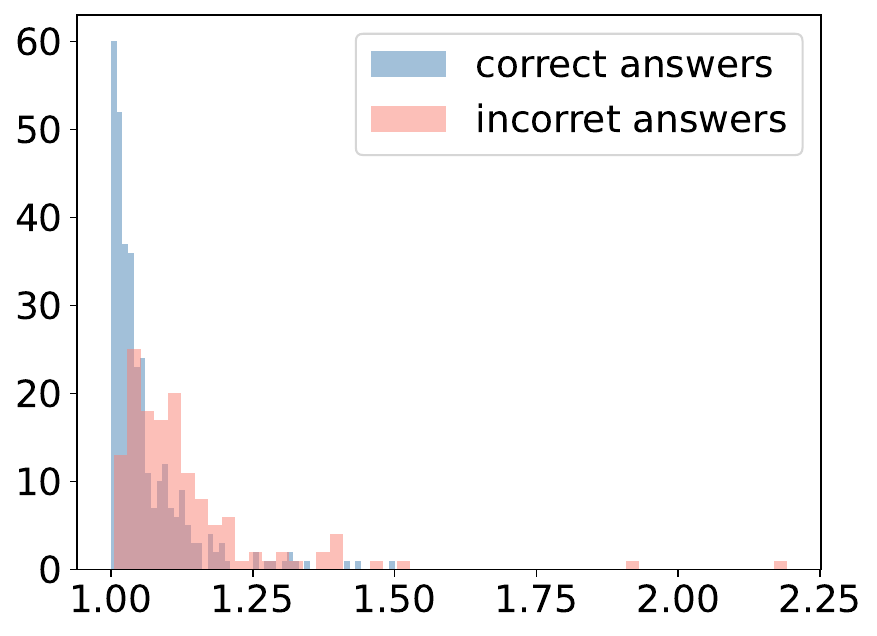}}
    \subfigure[POIE(OOD)]{\includegraphics[width=.24\textwidth]{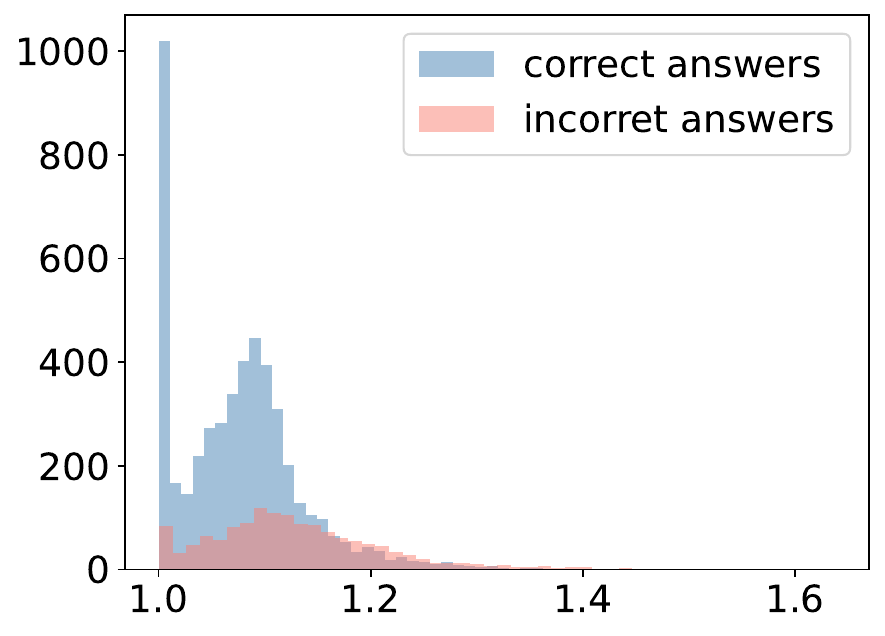}}
    \subfigure[CORD(OOD)]{\includegraphics[width=.24\textwidth]{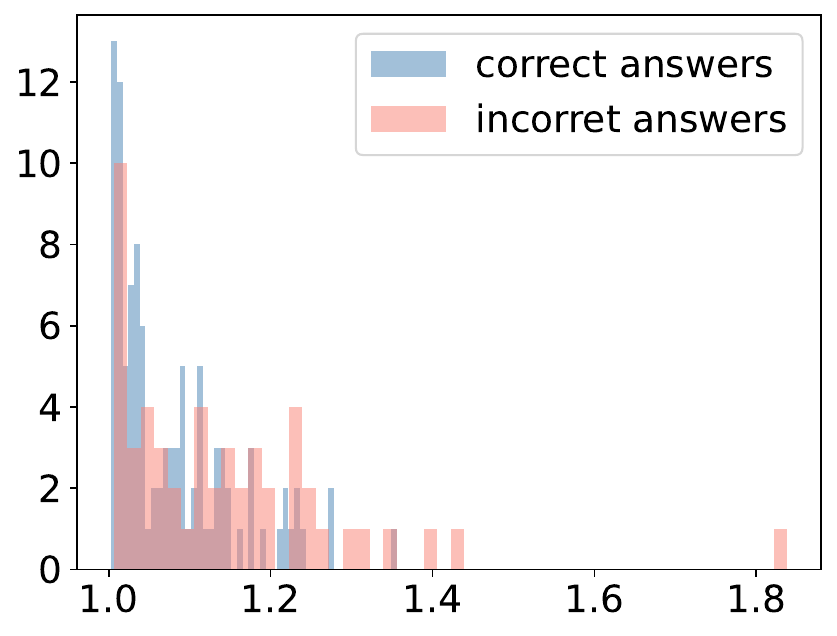}}
    \subfigure[SROIE]{\includegraphics[width=.24\textwidth]{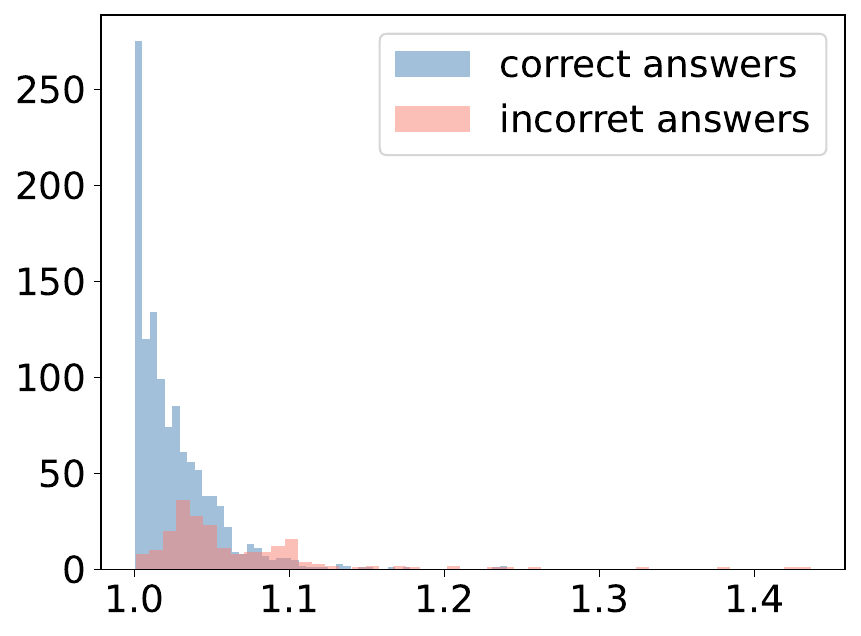}}
    \caption{PPL distribution of Correct and Incorrect Predictions. X-axis is PPL, Y-axis is frequency.}
    \label{fig:good_bad_ppl_7b}
\end{figure*}

First, Figure~\ref{fig:acc_vs_ppl_7b} shows an inverse relationship between overall accuracy and mean PPL scores at the dataset level, indicating that lower confidence (i.e., higher PPL scores) correlates with lower prediction accuracy. Further analysis in Figure~\ref{fig:good_bad_ppl_7b} and Table~\ref{tab:good_vs_bad_mean_ppl_7b} compares correct and incorrect predictions, revealing that correct answers have significantly lower PPL scores than incorrect ones, with a clear distribution difference.

Additionally, Table~\ref{tab:infer_ppl_7b} demonstrates that thresholding PPL improves performance in most datasets compared to either complete short-answer or long-form reasoning inference. These experimental results confirm that PPL scores serve as a reliable indicator of the correctness of LLM-generated short answers.

\begin{table}[t]
\centering
\scriptsize
\begin{tabular}{l|cccccccc|c}
\toprule
 \textbf{PPL} &  \textbf{DocVQA} & \textbf{ChartQA} & \textbf{InfoVQA} & \textbf{VisualMRC} & \textbf{FUNSD} & \textbf{POIE} & \textbf{CORD} & \textbf{SROIE} & \textbf{Average} \\
\midrule
Correct Cases & 1.09 & 1.09 & 1.12 & 1.08 & 1.05 & 1.07 & 1.07 & 1.02 & 1.07  \\
Incorrect Cases & 1.21 & 1.24 & 1.23 & 1.13 & 1.13 & 1.14 & 1.14 & 1.07 & 1.17 \\
\bottomrule
\end{tabular}
\caption{Mean PPL scores for correct and incorrect cases across datasets.}\label{tab:good_vs_bad_mean_ppl_7b}
\end{table}

\begin{table}[!t]
\centering
\small
\begin{tabular}{l|cccccccccc}
\toprule
 \textbf{Method} &  \textbf{DocVQA} & \textbf{ChartQA} & \textbf{InfoVQA} & \textbf{VisualMRC} & \textbf{FUNSD} & \textbf{POIE} & \textbf{CORD} & \textbf{SROIE}  \\
\midrule
Short & 68.7& 35.5 & \textbf{35.8}  & 22.3 & 70.0 & 77.1 &66.4 & 84.8  \\
Long & 67.1 & 30.3 & 34.0 &  \textbf{25.0} & 59.3 &72.2 & 58.7 & \textbf{88.0}   \\
PPL & \textbf{72.1} & \textbf{35.6} & 35.5 & 22.8 & \textbf{70.1} & \textbf{77.5} & \textbf{67.8} & 87.3  \\
\bottomrule
\end{tabular}
\caption{Performance of different inference methods across datasets.}\label{tab:infer_ppl_7b}
\end{table}

\section{Hybrid CAR with TALE}\label{app:combined_method}

This section explores the feasibility of integrating CAR with token reduction techniques such as TALE~\cite{han2024token}. Most reasoning token reduction methods operate by replacing the ground-truth reasoning process with a shorter, LLM-generated version, which makes it possible to combine with CAR. In this section, we combine CAR with TALE by substituting the original reasoning steps with TALE-generated responses.

We evaluate this approach using Qwen2.5-7B and Llama3.1-8B, resulting in two variants: \textbf{CAR-TALE\textsubscript{Qwen2.5}} and \textbf{CAR-TALE\textsubscript{Llama3.1}}. Each variant further supports short- and long-form reasoning, yielding another four variants: \textbf{CAR-TALE\textsubscript{Qwen2.5-Short}}, \textbf{CAR-TALE\textsubscript{Qwen2.5-Long}}, \textbf{CAR-TALE\textsubscript{Llama3.1-Short}}, \textbf{CAR-TALE\textsubscript{Llama3.1-Long}}. 

The experimental results, presented in Table~\ref{tab:exp_qwen_car_tale} and Table~\ref{tab:exp_llama_tale_car}, demonstrate the effectiveness of combining CAR with TALE for reasoning token reduction.

When using Qwen2.5, CAR achieves a significant performance improvement of 6.7\% (85.5\% vs. 78.8\%) when integrated with TALE, while also reducing the number of generated tokens (111.3 vs. 127.8). A similar result is observed with Llama3.1, where CAR's performance improves notably (80.8\% vs. 71.6\%) when paired with TALE.

These results highlight the synergistic benefits of fusing CAR with other reasoning token reduction methods, further validating their combined effectiveness.

\begin{table*}[t]
\centering
\small
\begin{tabular}{l|cc}
\toprule
~ & \multicolumn{2}{c}{\textbf{GSM8K}}  \\  \midrule
\textbf{Model} & Acc & \#Token  \\
\midrule
\textbf{Baselines} &  &    \\
\midrule
% \rowcolor{gray!20} Upper Bound & 92.4 & - & 75.2 & - & 76.0 & - & - & - \\
Qwen2.5\textsubscript{Short}~\cite{yang2024qwen2} & 24.2 & 9.2  \\
Qwen2.5\textsubscript{Long}~\cite{yang2024qwen2} & 76.4 & 138.4  \\
\midrule
\textbf{SOTA} &  &    \\
\midrule
% \rowcolor{gray!20} Upper Bound & 92.4 & - & 75.2 & - & 76.0 & - & - & - \\
TALE\textsubscript{Qwen2.5}~\cite{han2024token} & \textbf{87.6} & 124.0  \\
\midrule
\textbf{CAR} &  &    \\
\midrule
% \rowcolor{gray!20}
% CAR\textsubscript{Qwen2.5-UpperBound} & 80.1 & 115.0 & 89.0 & 18.1 & 87.4 & 47.4 & 85.5 & 60.2 \\
CAR\textsubscript{Qwen2.5-Short} & 23.6 & 9.4  \\
CAR\textsubscript{Qwen2.5-Long} & 76.8 & 139.4  \\
CAR\textsubscript{Qwen2.5} & 78.8 & 127.8  \\
\midrule
\textbf{CAR-TALE} &  &   \\
\midrule
% \rowcolor{gray!20}
% CAR\textsubscript{Qwen2.5-UpperBound} & 80.1 & 115.0 & 89.0 & 18.1 & 87.4 & 47.4 & 85.5 & 60.2 \\
CAR-TALE\textsubscript{Qwen2.5-Short}  & 23.9 & 9.1\\
CAR-TALE\textsubscript{Qwen2.5-Long}  & 86.5 & 116.9 \\
CAR-TALE\textsubscript{Qwen2.5}  &  85.5 & 111.3\\
% Qwen2.5-0.5B\textsubscript{Short}                        &        \textbf{56.83}        &         \textbf{31.92}         &         \textbf{58.02}        &   \textbf{48.92} \\
% Qwen2.5-0.5B\textsubscript{Long}                       &           44.01        &   23.52      &            44.53   &  37.35
\bottomrule
\end{tabular}%
\caption{Performance of CAR fused with TALE using Qwen2.5.}
\label{tab:exp_qwen_car_tale}
\end{table*}

\begin{table*}[t]
\centering
\small
\begin{tabular}{l|cc}
\toprule
~ & \multicolumn{2}{c}{\textbf{GSM8K}}  \\  \midrule
\textbf{Model} & Acc & \#Token  \\\midrule
\textbf{Baselines} &  &   \\
\midrule
% \rowcolor{gray!20} Upper Bound & 92.4 & - & 75.2 & - & 76.0 & - & - & - \\
Llama3.1\textsubscript{Short}~\cite{grattafiori2024llama} & 23.9 & 9.1 \\
Llama3.1\textsubscript{Long}~\cite{grattafiori2024llama} & 72.9 & 118.9  \\
\midrule
\textbf{SOTA} &  &   \\
\midrule
% \rowcolor{gray!20} Upper Bound & 92.4 & - & 75.2 & - & 76.0 & - & - & - \\
TALE\textsubscript{Llama3.1}~\cite{han2024token} & 80.6 & 183.2  \\
\midrule
% \rowcolor{gray!20}
% CAR\textsubscript{Llama3.1-UpperBound} & 77.4 & 99.4 & 88.3 & 40.4 & 81.9 & 102.7 & 82.5 & 80.8 \\
\textbf{CAR} &  &   \\
\midrule
CAR\textsubscript{Llama3.1-Short} & 21.3 & 9.1  \\
CAR\textsubscript{Llama3.1-Long} & 73.5 & 116.2  \\
CAR\textsubscript{Llama3.1} & 71.6 & 108.9  \\\midrule
\textbf{CAR-TALE} &  &    \\
\midrule
CAR-TALE\textsubscript{Llama3.1-Short} & 22.5 & 9.1  \\
CAR-TALE\textsubscript{Llama3.1-Long} & \textbf{81.5} & 183.5   \\
CAR-TALE\textsubscript{Llama3.1} & 80.8 & 173.2  \\
% Qwen2.5-0.5B\textsubscript{Short}                        &        \textbf{56.83}        &         \textbf{31.92}         &         \textbf{58.02}        &   \textbf{48.92} \\
% Qwen2.5-0.5B\textsubscript{Long}                       &           44.01        &   23.52      &            44.53   &  37.35
\bottomrule
\end{tabular}%

\caption{Performance of CAR fused with TALE using Llama3.1.}
\label{tab:exp_llama_tale_car}
\end{table*}

\end{document}